\documentclass[lettersize,journal]{IEEEtran}
\usepackage{amsmath,amsfonts}
\usepackage{algorithmic}
\usepackage{algorithm}
\usepackage{array}
\usepackage[caption=false,font=normalsize,labelfont=sf,textfont=sf]{subfig}
\usepackage{textcomp}
\usepackage{stfloats}
\usepackage{url}
\usepackage{verbatim}
\usepackage{graphicx}
\usepackage{cite}

\usepackage{xcolor}
\definecolor{link}{HTML}{73140C}  
\definecolor{cite}{HTML}{000073}  
\definecolor{url}{HTML}{0B03C5} 

\usepackage[colorlinks=true, linkcolor=link, citecolor=cite, urlcolor=url]{hyperref}
\usepackage[utf8]{inputenc}        
\usepackage[T1]{fontenc}           
\usepackage{nicefrac}              
\usepackage{microtype}             
\usepackage{makecell}
\usepackage{multirow}
\usepackage[table,xcdraw]{xcolor}
\usepackage{booktabs}
\usepackage[capitalize,noabbrev]{cleveref}
\usepackage{colortbl}
\usepackage{pifont}
\usepackage[normalem]{ulem}
\usepackage{wrapfig}
\usepackage[many]{tcolorbox}
\usepackage{listings}
\usepackage{arydshln}

\newcommand{\cmark}{\checkmark}   
\usepackage{amssymb}

\crefname{figure}{Fig.}{Figs.}

\begin{document}

\title{MVEI \& EmObserver: Empowering MLLM-Oriented Visual Emotional Intelligence via Emotion Statement Judgement}


\author{Daiqing Wu, Dongbao Yang, Jiashu Yao, Hongrui Zhang, Can Ma, Yu Zhou, and Sicheng Zhao,~\IEEEmembership{Senior Member,~IEEE}%
\thanks{Daiqing Wu, Hongrui Zhang, and Sicheng Zhao are with the Department of Psychological and Cognitive Sciences, Tsinghua University, Beijing 100084, China. E-mail: wudaiqing04@gmail.com, smilingweeping@gmail.com, schzhao@tsinghua.edu.cn.}
\thanks{Dongbao Yang and Yu Zhou are with Nankai University, Tianjin 300071, China. E-mail: yangdongbao@nankai.edu.cn, yzhou@nankai.edu.cn.}
\thanks{Jiashu Yao is with Beijing Institute of Technology, Beijing 100081, China. E-mail: yaojiashu@bit.edu.cn.}
\thanks{Can Ma is with the Institute of Information Engineering, Chinese Academy of Sciences, Beijing 100085, China. E-mail: macan@iie.ac.cn.}
}


\markboth{MVEI \& EmObserver: Pre-Print Version}%
{Wu \MakeLowercase{\textit{et al.}}: MVEI \& EmObserver: Empowering MLLM-Oriented Visual Emotional Intelligence via Emotion Statement Judgement}


\maketitle

\begin{abstract}
Affective Image Content Analysis (AICA) aims to recognize and understand emotions elicited by visual content, representing an indispensable step toward Artificial General Intelligence (AGI). However, despite the rapid progress of Multimodal Large Language Models (MLLMs), systematic evaluation of their visual emotional intelligence remains largely absent from recent model releases. We attribute this gap to a structural mismatch between conventional AICA paradigms and the open-ended, instruction-driven nature of MLLMs, where further analysis reveals four major limitations: omission of plausible responses, limited emotion taxonomies, neglect of contextual factors, and labor-intensive annotation. To overcome these barriers, we introduce Emotion Statement Judgement (ESJ), a statement-verification formulation that preserves the expressiveness of the input space while constraining outputs to discriminative judgements. We further develop INSETS, a labor-efficient pipeline that instantiates ESJ at scale by constructing INSETS-462k and supporting MVEI, a rigorously refined benchmark spanning sentiment polarity, emotion interpretation, scene context, and perception subjectivity. Beyond evaluation, we build EmObserver, an emotion-oriented MLLM optimized on ESJ through an elaborate multi-stage recipe. Extensive evaluation of broad-spectrum MLLMs on MVEI reveals fine-grained insights into current artificial visual emotional intelligence, while experiments on multiple AICA benchmarks demonstrate the accuracy, generalization, and reasoning faithfulness of EmObserver. Collectively, these results establish ESJ as a practical formulation, MVEI as a comprehensive benchmark, and EmObserver as an advanced baseline for advancing MLLM-oriented visual emotional intelligence. Code will be released at: \href{https://github.com/wdqqdw/EmObserver}{https://github.com/wdqqdw/EmObserver}.
\end{abstract}

\begin{IEEEkeywords}
Visual Emotional Intelligence, Emotion Statement Judgement, Emotion-Oriented MLLMs.
\end{IEEEkeywords}

\section{Introduction}
\label{sec:intro}

The ability to perceive, understand, and respond to affective signals in visual stimuli constitutes a fundamental component of human emotional intelligence, supporting sound decision-making and effective communication \cite{schutte2001decision}. The demand to computationally model this ability has established Affective Image Content Analysis (AICA), which seeks to recognize and understand emotions elicited by visual content, as an important research direction \cite{tpami2022review, mer-llm}. Meanwhile, Multimodal Large Language Models (MLLMs) are increasingly regarded as a promising pathway toward Artificial General Intelligence (AGI), driven by their rapid progress and repeated advances beyond human expectations across diverse tasks \cite{nips2023instructblip}. However, despite the indispensability of visual emotional intelligence \cite{minsky_emotion}, recently released MLLMs are rarely accompanied by corresponding evaluations \cite{gpt5,qwen3.5,seed20pro}. We argue that this puzzling omission stems from the difficulty of smoothly transferring conventional AICA paradigms to MLLMs, leaving the field without a critical foothold for evaluating and optimizing MLLM-oriented visual emotional intelligence.

\begin{figure*}
    \centering
    \includegraphics[width=1\linewidth]{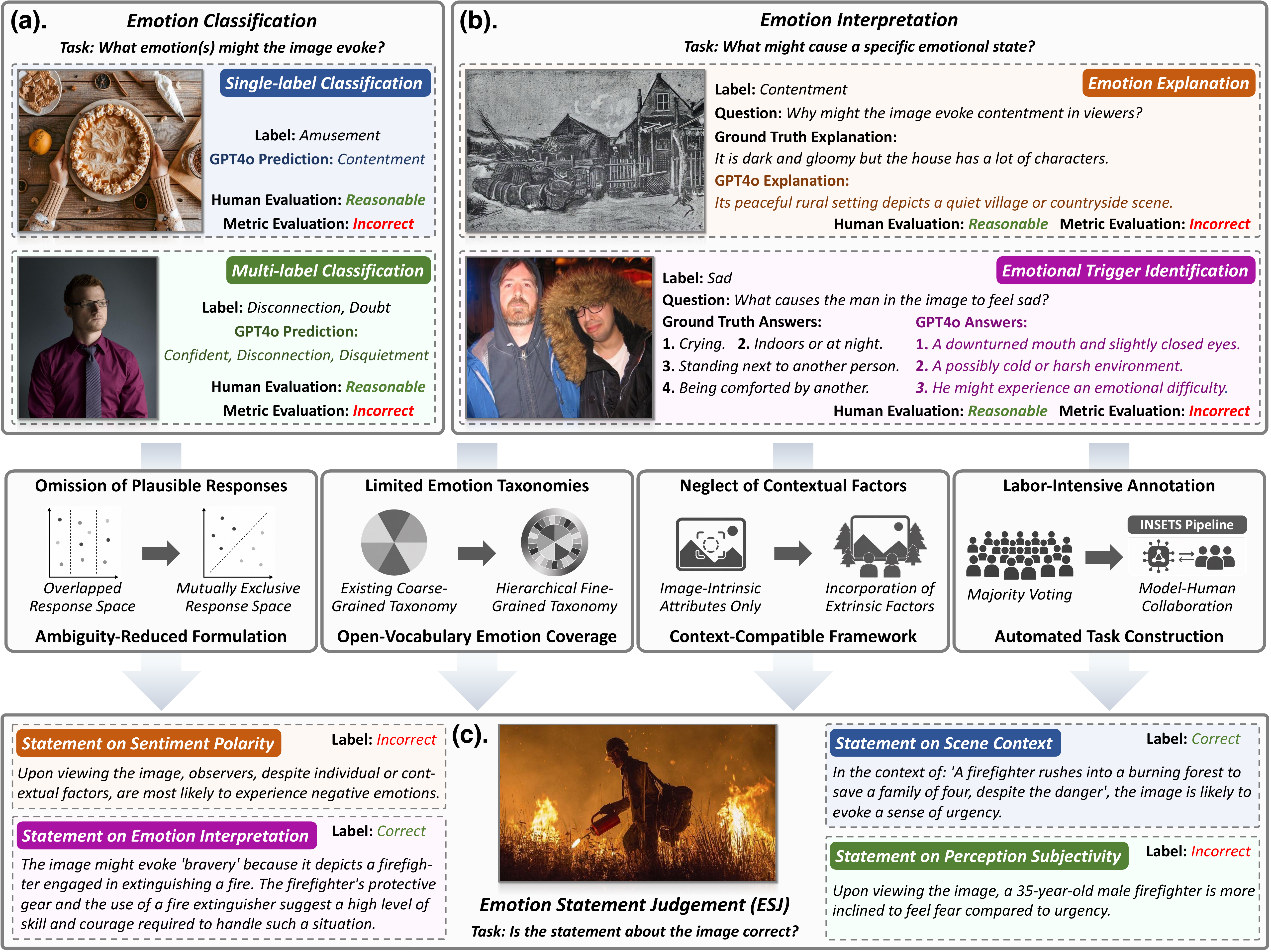}
    \vskip -0.1in
    \caption{\textbf{Comparison of prevailing AICA paradigms and the proposed ESJ formulation.} Conventional classification and interpretation approaches (a,b) are constrained by rigid answer spaces, limited affective and contextual coverage, and costly annotation, whereas ESJ (c), supported by INSETS, enables customized and scalable evaluation and optimization of MLLM-oriented visual emotional intelligence.}
    \label{fig:1}
    \vskip -0.10in
\end{figure*}

To pinpoint this mismatch, we revisit the dominant learning and evaluation practices in AICA, as summarized in \cref{fig:1}(a,b). Existing AICA tasks are commonly organized around two formulations: emotion classification and emotion interpretation. The former maps an image to one or more labels within a predefined taxonomy \cite{aaai2016fi,cvpr2017emotic,cvpr2020web,iccv2023emoset}, while the latter asks models to explain the rationale behind a specific interpretation \cite{cvpr2021artemis,cvpr2023affection,2024eibench}. Despite the diverse output forms, a similar pattern holds: models are trained on annotated data and then tested on held-out samples from the same answer space. This design offers clear supervision and standardized comparison, yet tightly couples model performance to a pre-specified task format. As a result, success on these tasks primarily reflects how well a model has learned task-specific emotional signals, rather than whether it possesses a broadly transferable capacity to reason about visual emotions without dedicated adaptation.

When applied to general-purpose MLLMs, this learning-and-evaluation paradigm exposes four major limitations. \textbf{\textit{First}}, fixed reference labels or answers define narrow response spaces that omit plausible alternatives. Visual emotion perception is inherently subjective \cite{mm2016opinion}, and the same image may elicit different yet reasonable responses. Consequently, restrictive supervision can discourage valid interpretations during learning, while reference-based metrics may reject them during evaluation, as illustrated by the GPT4o examples in \cref{fig:1} \cite{hurst2024gpt4o}. \textbf{\textit{Second}}, closed and coarse-grained emotion taxonomies constrain both what models can learn and what benchmarks can recognize. Widely used datasets such as FI \cite{aaai2016fi} and ArtEmis \cite{cvpr2021artemis} rely on only a small set of predefined emotions, collapsing fine-grained affective distinctions and leaving responses beyond their label spaces unsupported. \textbf{\textit{Third}}, existing paradigms mainly emphasize image-intrinsic attributes while underrepresenting contextual factors. Yet psychological studies show that emotional responses are also shaped by extravisual information \cite{barrett2011context}, including scene context \cite{wieser2012faces} and viewer-specific characteristics \cite{individual2004}, which limits systematic optimization and evaluation of context-conditioned affective reasoning. \textbf{\textit{Fourth}}, task-specific learning depends heavily on human-annotated emotion data, whose reliability often requires redundant crowdsourcing and majority voting \cite{cvpr2017crowdscource}. For instance, constructing ArtEmis required 6,788 crowd workers to provide 454,684 emotional responses and explanations, amounting to 11,138 hours of human labor \cite{cvpr2021artemis}. Such substantial annotation costs create a major bottleneck for scaling AICA data and extending evaluation benchmarks across domains.

To address these limitations and facilitate MLLM-oriented visual emotional intelligence, we introduce the Emotion Statement Judgement (\textbf{ESJ}) task. ESJ reformulates AICA as statement verification: the model judges whether an emotion-centric statement is appropriate for a given image. By preserving the expressiveness of natural-language affective descriptions in the input space, ESJ is flexible enough to cover fine-grained emotions, image-intrinsic cues, and extrinsic contextual factors. Meanwhile, it constrains the output space with a clear judgement target, reducing the potential ambiguity of open-ended responses, collectively providing a unified and tailored formulation for both supervised adaptation and standardized evaluation. To instantiate ESJ at scale, we further develop \textbf{INSETS} (\underline{\textbf{IN}}telligent Vi\underline{\textbf{S}}ual \underline{\textbf{E}}motion \underline{\textbf{T}}agger and \underline{\textbf{S}}tatement Constructor), a scalable annotation pipeline that assigns each image multiple open-vocabulary emotion labels and converts them into multifaceted emotion-centric statements. Through model-human collaboration with lightweight human labor, INSETS enables efficient construction of a diverse ESJ corpus.

Using INSETS, we first construct \textbf{INSETS-462k}, a large-scale ESJ corpus containing 462,369 emotion-centric statements over 17,716 images. Grounded in established theories of affective cognition, it covers four complementary dimensions: sentiment polarity \cite{1980circumplex}, emotion interpretation \cite{1971constants}, scene context \cite{barrett2011context}, and perception subjectivity \cite{individual2004}, as exemplified in \cref{fig:1}(c). To provide a reliable evaluation standard, we further derive \textbf{MVEI} (\underline{\textbf{M}}ultifaceted Evaluation of \underline{\textbf{V}}isual \underline{\textbf{E}}motional \underline{\textbf{I}}ntelligence) from this corpus through dedicated human refinement. Specifically, five annotators verify the sampled image-statement pairs with a Fleiss' kappa of 0.61, after which inconsistent or erroneous samples are manually revised or discarded. Consequently, the resulting MVEI benchmark contains 3,086 rigorously verified pairs for comprehensively evaluating the visual emotional intelligence of MLLMs.

Beyond MVEI, we also introduce \textbf{EmObserver}, an emotion-oriented MLLM optimized primarily under the ESJ formulation. Its training follows four progressive stages: Cold-Start Initialization, Sampling Sharpening, Error-Guided Refinement, and Analysis Reinforcement. By combining Group Relative Policy Optimization (GRPO) \cite{shao2024deepseekmath} with On-Policy Self-Distillation (OPSD) \cite{opsd}, this multi-stage recipe fully exploits the verifiable output space and expressive input space of ESJ, enabling the model to develop transferable affective reasoning and produce accurate, faithful answers across diverse AICA tasks.

Leveraging MVEI together with two well-established AICA benchmarks, EEmo-Bench \cite{mm25eemobench} and VECBench \cite{emocaliber}, we conduct extensive experiments covering 24 open-source MLLMs, 4 proprietary MLLMs, and 4 emotion-oriented MLLMs. The results reveal pronounced capability variations in current MLLMs, demonstrating that visual emotional intelligence does not consistently scale with model size or recency. Under the same evaluation protocol, EmObserver consistently surpasses competitors and generalizes beyond ESJ to tasks with different input formats and output requirements. Further diagnostic and qualitative analyses also highlight fine-grained benefits of the ESJ formulation and EmObserver. In summary, the contributions of this paper are fourfold:

\begin{itemize}
    \item We identify the incompatibility of conventional AICA paradigms for MLLMs and introduce the ESJ task as a customized formulation for MLLM-oriented visual emotional intelligence.
    \item Complementing ESJ, we develop the INSETS pipeline, which enables scalable annotation of open-vocabulary emotion labels and construction of multifaceted emotion-centric statements. Through labor-efficient human refinement, a high-quality MVEI benchmark is further curated.
    \item By optimizing on the ESJ task, we introduce EmObserver, an emotion-oriented MLLM capable of performing discerning and generalizable affective reasoning.
    \item By conducting in-depth evaluations on MVEI and related datasets across open-source, proprietary, and emotion-oriented MLLMs, we reveal their developmental trends, capability variations, and future potential. These results establish MVEI as a meaningful reference and EmObserver as a strong baseline, jointly advancing visual emotional intelligence in MLLMs.
\end{itemize}

An earlier version of this work appeared at ICLR 2026 \cite{iclr2026mvei}. The conference paper introduced the ESJ formulation, the INSETS pipeline, and the MVEI benchmark, along with initial evaluations of MLLMs and lightweight adaptation studies. This journal article substantially extends that work in three directions. First, it introduces EmObserver and a four-stage optimization framework for improving visual emotional intelligence in MLLMs. Second, it broadens the empirical coverage to recent MLLMs of different scales and purposes, and evaluates cross-benchmark generalization on EEmo-Bench and VECBench. Third, it adds stage-wise ablations, diagnostic controls, error analyses, temporal comparisons, and qualitative studies to systematically examine current model capabilities and the effects of ESJ-oriented optimization. These extensions expand the conference work from customized evaluation to a unified framework for evaluating and improving visual emotional intelligence in MLLMs.

\section{Related Work}
\subsection{Conventional AICA Paradigms}
AICA goes beyond perceptual semantics toward cognitive-level affective comprehension, constituting an essential aspect of AGI \cite{agi2020condition,tpami2022review}. To computationally model emotion, prior studies commonly draw on two principal psychological frameworks: Categorical Emotion Space (CES), which discretizes affective states into predefined emotion categories, and Dimensional Emotion Space (DES), which represents emotions as coordinates in continuous affective dimensions. These two frameworks offer complementary advantages. CES provides a language-aligned representation that is more interpretable and relatively easy to annotate, whereas DES offers greater flexibility and stronger sensitivity to fine-grained affective variations. Representative AICA datasets built upon CES include early resources such as IAPS \cite{mikels2005emotional} and Abstract \cite{mm2010abstract}, as well as larger-scale datasets with richer attribute annotations, such as FI \cite{aaai2016fi}, ArtEmis \cite{cvpr2021artemis}, EmoSet \cite{iccv2023emoset} and EIBench \cite{2024eibench}. Other studies construct AICA datasets based on DES or incorporate both frameworks, such as IESN \cite{mm2016opinion}, EMOTIC \cite{cvpr2017emotic}, and LUCFER \cite{wacv2019lucfer}.

Building on these foundations, earlier task-specific AICA studies mainly focused on designing specialized network architectures \cite{wscnet,eccv2022s2ver}, attention mechanisms \cite{cvpr2022mdan,padnet}, and optimization objectives \cite{artemisv2,cvpr2023prob,mm24pacl} to enhance the representational capacity of expert models for affective cues. With adequate supervision, these methods often achieve strong performance under predefined affective taxonomies. However, this paradigm faces a practical challenge in real-world applications, where emotion categories often need to be adapted, expanded, or redefined, usually entailing costly re-annotation and model retraining, thereby imposing a heavy burden on both data construction and model development. To mitigate this rigidity, zero-shot AICA studies \cite{iccv2019zero-shot1,mmasia2022zero-shot2} replace one-hot emotion labels with semantic embeddings, allowing models trained on seen categories to generalize to unseen ones. Nevertheless, this flexibility often comes with substantial performance degradation, making it difficult to achieve a desirable balance between generalizability and accuracy.

\subsection{AICA with MLLMs}
In recent years, the rapid progress of MLLMs has offered a promising route to alleviating this dilemma. Through large-scale training on multimodal data, MLLMs acquire strong instruction-following abilities and broad knowledge coverage, enabling them to adapt to diverse downstream AICA tasks through prompt reformulation while maintaining a high degree of task competence \cite{icml2025icl}. However, it remains noteworthy that, despite being an indispensable component of AGI, related AICA tasks are rarely included in the technical reports accompanying general-purpose MLLM releases. We attribute this omission largely to the mismatch between existing AICA datasets and the open-ended, instruction-driven evaluation paradigm of MLLMs.

This gap has inspired several pioneering MLLM-oriented efforts \cite{arxiv2023mmbigbench,eccv2024faba,hu2025emobench} to compensate for the incompatible evaluation. For example, EEmo-Bench \cite{mm25eemobench} formulates a multi-task evaluation suite covering perception, description, ranking, and assessment, and further introduces image-pair tasks for joint emotion understanding. VECBench \cite{emocaliber} systematically organizes different types of conventional AICA datasets and constructs the high-quality structured reasoning dataset VEC-CoT to support future research. Meanwhile, another line of work designs emotion-centric tasks to fine-tune MLLMs into emotion-oriented MLLMs, with representative examples including EmoVIT \cite{cvpr2024emovit}, Emotion-Qwen \cite{emotion-qwen}, EmoCaliber \cite{emocaliber}, and EEmo-Logic \cite{eemologic}. Despite their valuable insights, these studies leave two key issues insufficiently addressed. \textit{\textbf{First}}, their evaluation protocols remain closely tied to conventional formulations, such as fixed-category classification, semantic matching of textual descriptions, and Valence-Arousal-Dominance (VAD) regression, and therefore do not fully mitigate the four limitations identified in our previous analysis. \textit{\textbf{Second}}, the resulting models often exhibit limited generalization, achieving strong performance under their reported settings while remaining less competitive under alternative criteria.

\begin{figure*}[t]
    \centering
    \includegraphics[width=1\linewidth]{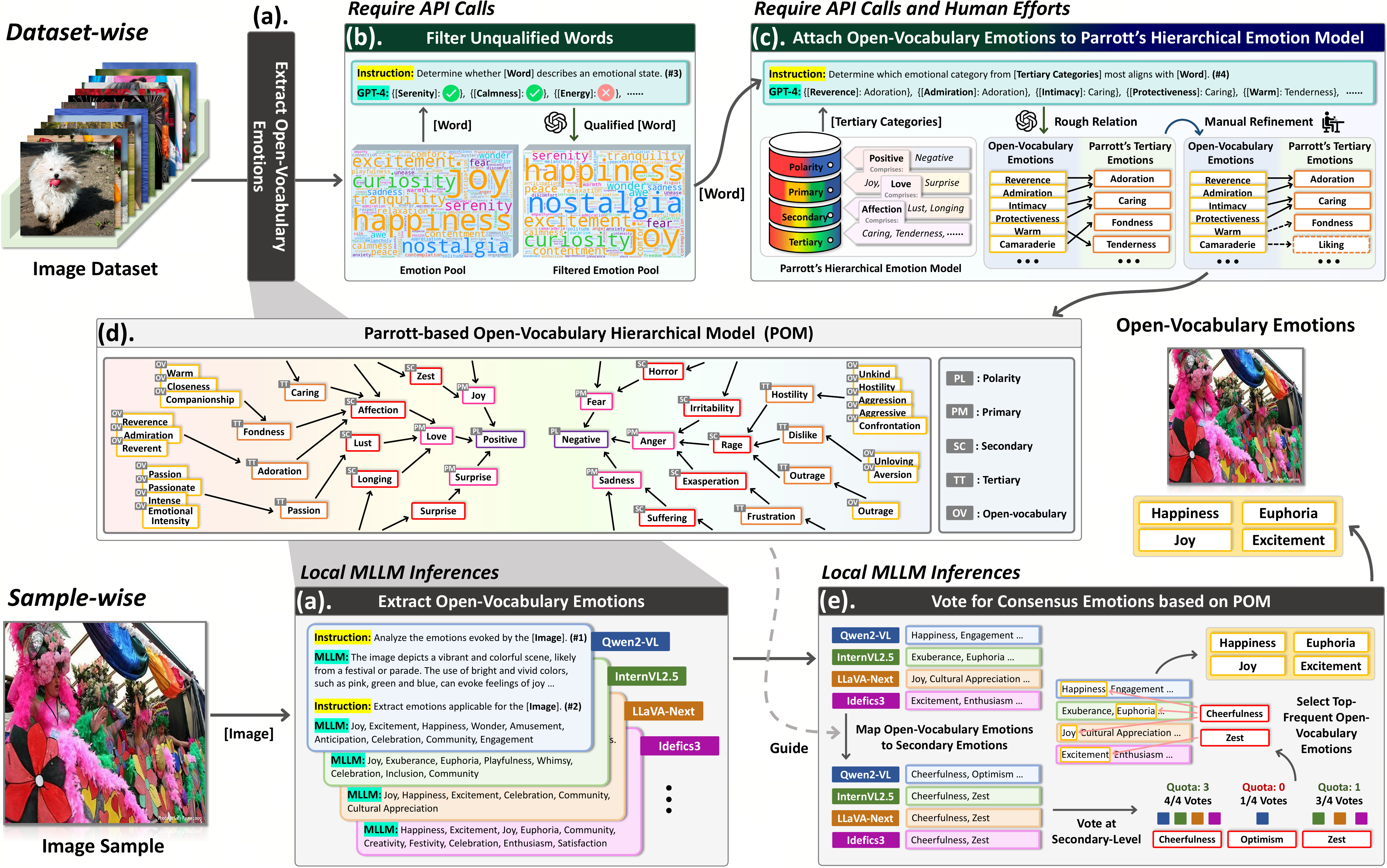}
    \vskip -0.10in
    \caption{\textbf{Overview of the open-vocabulary emotion tagging pipeline.} First, multiple local MLLMs extract candidate open-vocabulary emotions from raw images (a). At the dataset level, unqualified terms are filtered out (b), and the remaining emotions are mapped onto Parrott’s hierarchical emotion model and manually refined (c), yielding the Parrott-based Open-vocabulary Hierarchical Model (POM) shown in (d). Guided by POM, the image-level predictions from different MLLMs are mapped to secondary emotion categories and aggregated through consensus voting (e). The most frequently supported open-vocabulary emotions are then selected as the final labels for each image.}
    \vskip -0.10in
    \label{fig:2}
\end{figure*}

Motivated by these gaps, we propose ESJ as an MLLM-oriented task and construct MVEI to measure visual emotional intelligence beyond conventional fixed-form settings. We further introduce EmObserver, an emotion-oriented MLLM trained primarily under the ESJ formulation, which shows strong generalization not only on MVEI but also across alternative settings. Together, ESJ, MVEI, and EmObserver provide a unified basis for evaluating and improving AICA with MLLMs.

\section{The Emotion Statement Judgement Task}
\label{section:3}

The ESJ task is designed to evaluate whether MLLMs can make reliable affective judgements about visual content under a controlled yet expressive formulation. As illustrated in \cref{fig:1}(c), given an image and an emotion-centric statement, the model is required to determine whether the statement is correct with respect to the image. This formulation preserves the flexibility of natural-language descriptions while providing a clear target, making it suitable for both in-depth supervision and ambiguity-reduced evaluation. In our implementation, the prompt template used for ESJ is shown below:


\definecolor{PromptBlue}{RGB}{232,246,255}
\definecolor{PromptGray}{RGB}{245,245,245}
\definecolor{PromptBorder}{RGB}{120,160,190}

\begin{tcolorbox}[
    colback=PromptBlue,
    colframe=PromptBorder,
    boxrule=0.7pt,
    arc=2pt,
    left=2pt,
    right=2pt,
    top=2pt,
    bottom=2pt,
    title=\textbf{ESJ Prompt},
    coltitle=black,
    colbacktitle=PromptGray,
    fonttitle=\small\bfseries
]
\footnotesize
\texttt{\textless image\textgreater} Is the following statement correct about the image? \textbf{[STATEMENT]}. 
Choose the answer from \{`A': `Correct.', `B': `Incorrect.'\}. 
Answer in the format of \texttt{\textless think\textgreater...\textless/think\textgreater\textless answer\textgreater...\textless/answer\textgreater}.
\end{tcolorbox}

To provide a comprehensive assessment of visual emotional intelligence, ESJ organizes emotion-centric statements into four complementary dimensions inspired by cognitive studies \cite{2017cog+emo} of emotion and prior AICA research \cite{pieee2023label-effic}.

\begin{figure*}
    \centering
    \includegraphics[width=1\linewidth]{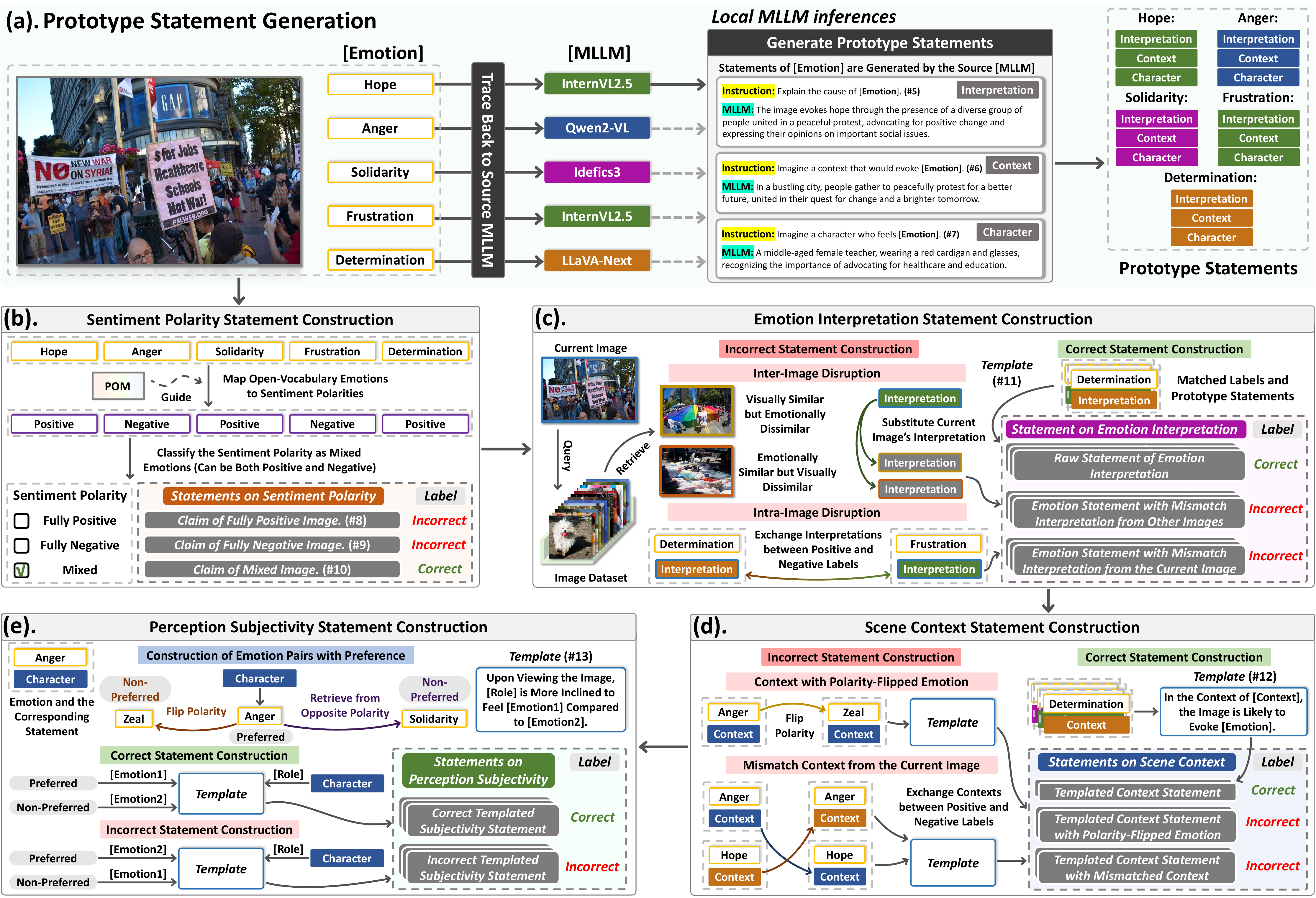}
    \vskip -0.10in
    \caption{\textbf{Overview of the emotion statement construction pipeline.} INSETS first traces each selected emotion label back to its source MLLM and generates prototype interpretation, context, and character statements (a). It then constructs ESJ statements from four dimensions: sentiment polarity (b), emotion interpretation (c), scene context (d), and perception subjectivity (e), where correct statements preserve matched label-prototype relations and incorrect statements are produced through controlled perturbations.}
    \label{fig:3}
    \vskip -0.1in
\end{figure*}

\textbf{1) \textit{Sentiment Polarity Statements}} evaluate whether MLLMs can identify the overall valence elicited by an image without providing contextual cues. They aim to assess MLLMs' proficiency in identifying the basic emotional tone.

\textbf{2) \textit{Emotion Interpretation Statements}} examine whether MLLMs can verify the consistency between an emotional state and its underlying explanation. This dimension goes beyond recognizing a label and instead probes whether the model can reason about why a particular emotion may be evoked.

\textbf{3) \textit{Scene Context Statements}} assess the model's ability to incorporate plausible extrinsic context when judging emotional responses. Rather than relying only on visible image-intrinsic attributes, these statements test whether MLLMs can connect the depicted content with broader scene-level situations that may shape affective interpretation.

\textbf{4) \textit{Perception Subjectivity Statements}} investigate whether MLLMs can account for viewer-dependent emotional responses. These statements introduce assumptions about observer identity, background, or perspective, requiring the model to judge how subjective factors may influence emotional perception.

Together, these four dimensions extend ESJ beyond conventional category prediction by covering both image-intrinsic affective cues and contextual factors that are critical to human emotional perception. As a result, ESJ provides a unified and extensible task formulation for evaluating the visual emotional intelligence of MLLMs.

\section{The INSETS Annotation Pipeline}

Complementing the ESJ task, we design INSETS, an automated pipeline for constructing emotion-centric statements at scale. INSETS consists of two major stages: open-vocabulary emotion tagging and emotion statement construction, corresponding to \cref{fig:2} and \cref{fig:3}, respectively. The former assigns diverse emotion labels to each image, while the latter converts these labels into verifiable emotion statements.

\begin{table*}[h]
\centering
\caption{Prompts and statement templates employed in the INSETS pipeline.}
\vskip -0.1in
\label{tab:1}
\footnotesize
\setlength{\tabcolsep}{5pt}
\renewcommand{\arraystretch}{1.08}
\resizebox{\linewidth}{!}{
\rowcolors{2}{white}{gray!8}
\begin{tabular}{p{0.055\linewidth}p{0.915\linewidth}}
\toprule
\textbf{ID} & \multicolumn{1}{c}{\textbf{Prompts and Statement Templates}} \\
\midrule
\#1 &
You are an Emotional Perception Expert. Please analyze the emotions that might be evoked by the given image.
Your analysis should explore a wide range of visual attributes, such as brightness, colorfulness, depicted scenes,
objects, human actions, and facial expressions. Additionally, provide detailed explanations linking these attributes
to the emotions they may trigger. If applicable, discuss any potential cultural or psychological factors influencing
these emotional responses. \\

\#2 &
You are an Emotional Perception Expert. Your task is to extract all applicable emotions as comprehensively as possible
based on the image description. Focus on distinct emotions such as happiness, sadness, fear, anger, etc.
Keep the list concise, with a maximum of 10 distinct emotions. \\

\#3 &
You are tasked with determining whether the word ``\textbf{[word]}'' describes a specific emotional state.
An emotional state is a psychological condition involving feelings and reactions triggered by internal or external events.
Respond with ``Yes'' if the word aligns with this definition, or ``No'' otherwise.
The output format should be \{``word'': ``response''\}. \\

\#4 &
You are tasked with assigning the word ``\textbf{[word]}'' to the most closely related emotional category from the following
113 predefined options: ``\textbf{[categories]}''. Consider broader semantic connections and possible emotional nuances
when making your judgement. If the word cannot reasonably fit any category, respond with ``not applicable''.
Do not create or assign new categories outside of the provided list. Do not provide any explanations or reasons for your choice.
The output format should be \{``word'': ``response''\}. \\

\#5 &
Briefly explain why this image might evoke ``\textbf{[emotion]}'' in viewers, without mentioning any other emotions. \\

\#6 &
Imagine a background story for the image that would evoke a sense of ``\textbf{[emotion]}'' in viewers.
Respond in one sentence. Do not merely restate the visible content. \\

\#7 &
Imagine a character who would feel ``\textbf{[emotion]}'' when viewing this image.
Include details such as their age, gender, profession, and other relevant traits.
Describe the character in one concise sentence without further explanation. \\

\#8 &
Upon viewing this image, observers, despite various individual or contextual factors, are most likely to experience positive emotions. \\

\#9 &
Upon viewing this image, observers, despite various individual or contextual factors, are most likely to experience negative emotions. \\

\#10 &
Upon viewing this image, observers are equally likely to experience either positive or negative emotions,
depending on individual or contextual factors. \\

\#11 &
Therefore, the image might evoke ``\textbf{[emotion]}'' in viewers. \\

\#12 &
In the context of: ``\textbf{[context]}'', the image is likely to evoke a sense of ``\textbf{[emotion]}''. \\

\#13 &
Upon viewing the image, ``\textbf{[role]}'' is more inclined to feel ``\textbf{[emotion1]}'' compared to ``\textbf{[emotion2]}''. \\
\bottomrule
\end{tabular}
\rowcolors{2}{}{}
}
\vskip -0.1in
\end{table*}

\subsection{Open-vocabulary Emotion Tagging}

The open-vocabulary emotion tagging stage aims to assign each image a set of reliable emotion labels that can later support statement construction. Its overall procedure is shown in \cref{fig:2}. Although recent studies suggest that MLLMs can generate rich emotional descriptions from visual content and infer underlying emotions from these descriptions \cite{nips2024emollama}, such outputs may still suffer from hallucinations \cite{arxiv2024hallucination}, trustworthiness issues \cite{eccv2024safebench}, and incomplete affective perception. INSETS therefore adopts an ensemble-based strategy: multiple MLLMs independently propose candidate emotions, while a hierarchy-guided voting mechanism consolidates their outputs into consensus labels.

Given an image, we first prompt multiple MLLMs to analyze the image-evoked emotions separately (with prompt \#1 in \cref{tab:1}, which we will abbreviate as \#1 later) and extract applicable emotion words (\#2) (\cref{fig:2}(a)). Repeating this process over the entire image dataset yields a dataset-level emotion pool containing diverse open-vocabulary candidates. Since this raw pool includes unsuitable words, such as non-emotional concepts or overly vague descriptors, we perform lexical vetting with GPT-4 \cite{arxiv2023gpt4}. Specifically, GPT-4 is prompted to determine whether each candidate word denotes an emotional state (\#3), producing a filtered emotion pool (\cref{fig:2}(b)). We select GPT-4 for this step because of its superior linguistic competence in emotion understanding \cite{acl2024emobench}.

To impose a comparable structure on free-form candidates, we continue by anchoring them to the well-established Parrott's hierarchical emotion model \cite{parrott2001emotions} (\cref{fig:2}(c)). Parrott's model organizes emotions into a multi-level taxonomy with 6 primary, 25 secondary, and 113 tertiary categories, capturing hierarchical affective relations from coarse-grained polarities to fine-grained states. Intuitively, anchoring each candidate to this model allows open-vocabulary labels to inherit higher-level affective relations while retaining their fine-grained lexical meanings. In practice, GPT-4 is prompted to assign each retained emotion word to the closest tertiary category (\#4). Since some candidates may be semantically ambiguous, the GPT-derived mappings are considered preliminary and therefore further refined by a hired human expert. This process yields an extended taxonomy, termed the \underline{\textbf{P}}arrott-based \underline{\textbf{O}}pen-vocabulary Hierarchical \underline{\textbf{M}}odel (\textbf{POM}) (\cref{fig:2}(d)), which supports multi-level tracing of open-vocabulary emotions for subsequent procedures.

Finally, INSETS uses POM to select consensus emotion labels for each image, as shown in \cref{fig:2}(e). Candidate emotions extracted by different MLLMs are first mapped to their corresponding secondary categories in POM. Voting is then conducted at this level to allocate category-wise quotas, reducing the impact of nuanced lexical variation among semantically related emotion words. Within each selected secondary category, candidate open-vocabulary emotions are ranked by their occurrence frequency across MLLMs, and the top-ranked labels are retained. This hierarchy-aware voting procedure improves annotation reliability while preserving the granularity and flexibility of open-vocabulary emotion labels.

\subsection{Emotion Statement Construction}

Based on the assigned emotion labels, INSETS constructs automatically annotated emotion-centric statements for ESJ, as illustrated in \cref{fig:3}. The construction begins with the prototype statement generation stage (\cref{fig:3}(a)). For each emotion label, we trace it back to the source MLLM that originally extracted it, which is then prompted to generate three types of prototypes regarding the emotion: \textbf{1)} a \textit{prototype interpretation} that explains the rationale behind the evoked emotion (\#5); \textbf{2)} a \textit{prototype context} that describes a plausible background situation that may trigger the emotion (\#6); and \textbf{3)} a \textit{prototype character} that describes a hypothetical observer who may experience the emotion (\#7). This source-tracing design keeps each prototype semantically tied to the model that proposed the corresponding emotion label. From a dataset-level perspective, prototype generation is distributed across multiple MLLMs, introducing lexical and reasoning diversity into subsequent statement construction.

\textbf{\textit{Sentiment Polarity Statement Construction}} (\cref{fig:3}(b)):
For sentiment polarity, INSETS assigns each image to an applicable polarity class according to POM. We set three mutually exclusive polarity types: \textbf{1)} \textit{Fully Positive}, indicating all labels reside in the positive spectrum; \textbf{2)} \textit{Fully Negative}, reflecting scenarios where all labels fall within the negative spectrum; and \textbf{3)} \textit{Mixed}, if at least two positive and two negative labels are present. Next, three predefined statements (\#8, \#9, \#10) are instantiated for images with properly assigned polarities, and their respective labels are determined accordingly.

\textbf{\textit{Emotion Interpretation Statement Construction}} (\cref{fig:3}(c)):
Emotion interpretation statements are constructed by combining a prototype interpretation with an emotional conclusion (\#11). For correct statements, the interpretation and the target emotion are drawn from the matched source. For incorrect statements, we introduce two contrastive disruption strategies. The first is \textit{inter-image disruption}, where we retrieve two external images from the dataset: one that is visually similar but emotionally dissimilar, and another that is emotionally similar but visually dissimilar. The former tests whether MLLMs can detect affective discrepancies beyond visual resemblance \cite{spm2006gap}, while the latter examines whether they can identify the actual emotional trigger rather than relying on surface-level semantic similarity. Visual similarity is measured by CLIP \cite{icml2021clip}, and emotional similarity is obtained by comparing at the tertiary level of POM. The second strategy is \textit{intra-image disruption}, where interpretations are exchanged between opposite-polarity labels within the same image. This creates hard negatives that require models to establish precise causal links between visual evidence and specific emotional states.

\begin{table}[t]
\centering
\caption{Statistics of the MLLMs used in INSETS. Bold and underline marks indicate column-wise maximum and minimum values.}
\vskip -0.1in
\label{tab:2}
\small
\setlength{\tabcolsep}{4.2pt}
\resizebox{0.9\columnwidth}{!}{
\begin{tabular}{lccc}
\toprule
\multirow[c]{2}{*}{\textbf{Models}} & \multicolumn{3}{c}{Statistics on INSETS-462k} \\
\cmidrule(lr){2-4}
& \makebox[0.2\linewidth][c]{\makecell{Extracted\\Emotions}}
& \makebox[0.2\linewidth][c]{\makecell{Selected\\Emotions}}
& \makebox[0.2\linewidth][c]{\makecell{Generated\\Statements (\%)}} \\
\midrule
LLaVA-Next \cite{cvpr2024llava_next} {\color{gray}\scriptsize 7.6B} & 8.3 & 2.4 & 9.8 \\
Mantis \cite{arxiv2024mantis} {\color{gray}\scriptsize 8.5B} & \textbf{12.6} & \textbf{2.9} & \textbf{13.1} \\
mPLUG-Owl3 \cite{arxiv2024mplugowl3} {\color{gray}\scriptsize 8.1B} & 9.2 & 2.7 & 11.2 \\
Idefics3 \cite{arxiv2024idefics3} {\color{gray}\scriptsize 8.5B} & 10.0 & \textbf{2.9} & 12.5 \\
Phi-3.5-Vision \cite{arxiv2024phi3} {\color{gray}\scriptsize 4.1B} & 9.9 & 2.8 & 11.7 \\
Qwen2-VL \cite{arxiv2024qwen2vl} {\color{gray}\scriptsize 8.3B} & 8.8 & 2.7 & 10.9 \\
Llama-3.2-Vision \cite{arxiv2024llama3} {\color{gray}\scriptsize 10.7B} & \underline{7.2} & \underline{2.3} & \underline{9.3} \\
Molmo \cite{arxiv2024molmo} {\color{gray}\scriptsize 8.0B} & 10.8 & 2.7 & 12.0 \\
InternVL2.5 \cite{arxiv2024intervl25} {\color{gray}\scriptsize 8.3B} & 8.5 & \underline{2.3} & 9.5 \\
\bottomrule
\end{tabular}
}
\vskip -0.1in
\end{table}

\begin{table}[t]
\centering
\caption{Statistics of the human refinement process. SP, EI, SC, and PS denote sentiment polarity, emotion interpretation, scene context, and perception subjectivity.}
\label{tab:3}
\vskip -0.1in
\small
\setlength{\tabcolsep}{4.5pt}
\resizebox{0.9\linewidth}{!}{
\begin{tabular}{lccccc}
\toprule
\multirow[c]{2}{*}{\textbf{Items}}
& \multicolumn{5}{c}{Statistics on MVEI (\%)} \\
\cmidrule(lr){2-6}
& \makebox[0.15\linewidth][c]{SP} & \makebox[0.15\linewidth][c]{EI} & \makebox[0.15\linewidth][c]{SC} & \makebox[0.15\linewidth][c]{PS} & \makebox[0.15\linewidth][c]{Total} \\
\multicolumn{6}{c}{\cellcolor[HTML]{EBEBEB}Annotation Agreement} \\
5/5 & 61.0 & 42.5 & 78.1 & 44.0 & 54.0 \\
4/5 & 33.2 & 46.6 & 15.9 & 43.7 & 36.6 \\
3/5 & 1.3 & 1.3 & 1.3 & 1.7 & 1.4 \\
2/5 & 1.0 & 0.7 & 0.4 & 2.4 & 1.1 \\
1/5 & 2.5 & 3.6 & 1.9 & 3.9 & 3.1 \\
0/5 & 1.0 & 5.3 & 2.4 & 4.3 & 3.8 \\
\textit{Kappa} & 0.68 & 0.51 & 0.81 & 0.52 & 0.61 \\
\multicolumn{6}{c}{\cellcolor[HTML]{EBEBEB}Construction Accuracy} \\
\ding{51} Pairs & 94.9 & 86.2 & 94.6 & 87.5 & 89.7 \\
\ding{55} Pairs & 93.4 & 92.0 & 93.4 & 88.0 & 91.5 \\
\bottomrule
\end{tabular}
}
\vskip -0.1in
\end{table}

\begin{table}[t]
\centering
\caption{Overall statistics of INSETS-462k and MVEI.}
\vskip -0.1in
\label{tab:4}
\small
\setlength{\tabcolsep}{9pt}
\resizebox{0.8\linewidth}{!}{
\begin{tabular}{lcc}
\toprule
\textbf{Items} & INSETS-462k & MVEI \\
\midrule
Number of Images & 17,716 & 3,086 \\
Number of Statements & 462,369 & 3,086 \\
Emotion Labels Per Image & 4.9 & 5.2 \\
Distinct Emotion Labels & 751 & 424 \\
Statements Per Image & 26.1 & 1.0 \\
Average Length of Statements & 39.0 & 36.5 \\
\bottomrule
\end{tabular}
}
\vskip -0.1in
\end{table}

\textbf{\textit{Scene Context Statement Construction}} (\cref{fig:3}(d)):
Scene context statements are formed by pairing a prototype context with an emotional conclusion (\#12). Correct statements combine the context and emotion associated with the same label. For incorrect statements, INSETS applies two perturbation strategies. The first is a \textit{polarity-flip operation}, which replaces the original label with a tertiary emotion randomly sampled from the opposite polarity spectrum in POM. The second swaps prototype contexts between positive and negative labels within the same image. These strategies produce statements in which the contextual description remains plausible in isolation but becomes incompatible with the stated emotional conclusion, thereby testing whether MLLMs can reason about the interaction between scene context and image-evoked emotion.

\textbf{\textit{Perception Subjectivity Statement Construction}} (\cref{fig:3}(e)):
This proportion is generated by combining a prototype character with a preference relation between two candidate emotions (\#13). For each character, the preferred emotion is the one associated with its corresponding label, while the non-preferred emotion is obtained either from an opposite-polarity label within the same image or through polarity-flip sampling in POM. Correct statements preserve this canonical preference order, whereas incorrect statements are produced by reversing the preferred and non-preferred emotions. This construction explicitly introduces viewer-dependent assumptions and probes whether MLLMs can account for how observer identity, background, or perspective modulates emotional perception.

\begin{figure*}[t]
    \centering
    \includegraphics[width=0.94\linewidth]{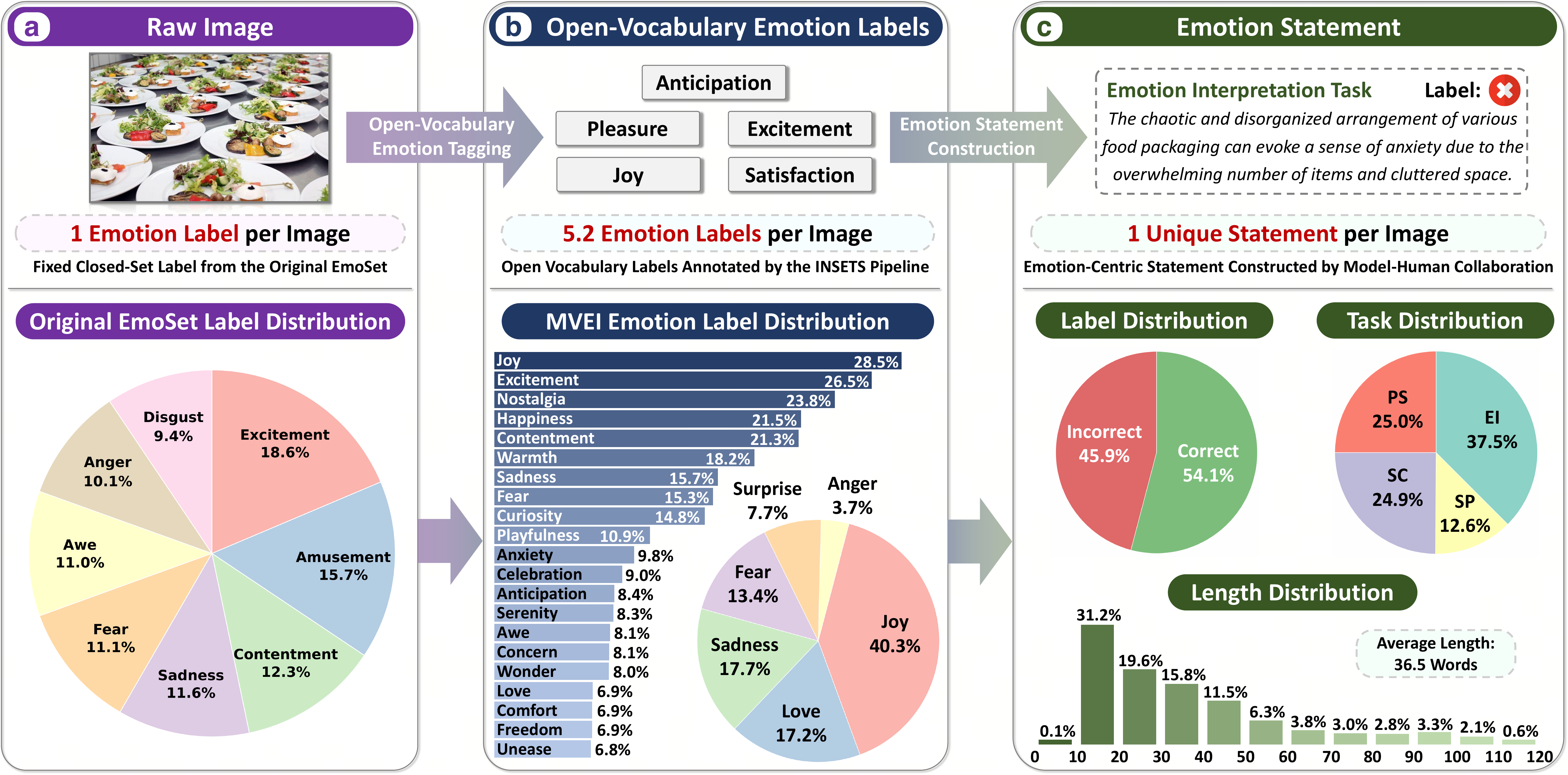}
    \vskip -0.10in
    \caption{\textbf{A closer look at MVEI.} The upper flow illustrates a representative sample, where a raw image is annotated with open-vocabulary emotion labels and converted into an emotion-centric statement. The lower parts show dataset statistics at different stages, including the original EmoSet label distribution, MVEI emotion label distributions at open-vocabulary and primary POM levels, and statement statistics covering label balance, task distribution, and length distribution.}
    \label{fig:4}
    \vskip -0.10in
\end{figure*}

\subsection{The INSETS-462k Dataset}

We instantiate the INSETS pipeline on EmoSet \cite{iccv2023emoset}, from which 17,716 images are selected as the visual source. Nine recent MLLMs with impressive performance \cite{2023opencompass} are employed for open-vocabulary emotion extraction and prototype statement generation, covering diverse model families and inference behaviors. Their contributions to the annotation process are summarized in \cref{tab:2}. As shown in the table, the extracted emotions, selected labels, and generated prototype statements are reasonably distributed across models, which helps avoid over-reliance on a single MLLM and introduces diversity into the constructed corpus. For the manual refinement of POM attachment, we hire a psychology postgraduate with formal training. This process takes approximately 15 hours.

Applying INSETS to these images produces INSETS-462k, a large-scale automatically annotated ESJ corpus. As reported in \cref{tab:4}, INSETS-462k contains 462,369 emotion-centric statements over 17,716 images, with an average of 4.9 emotion labels and 26.1 statements per image. It covers 751 distinct open-vocabulary emotion labels, substantially expanding the affective label space beyond closed-set taxonomies. 

\subsection{The MVEI Benchmark}

To obtain a reliable benchmark for evaluation, we further curate MVEI from INSETS-462k through human refinement. We sample 3,164 distinct image-statement pairs and recruit five graduate students to verify the correctness of the automatically assigned labels under detailed annotation guidelines. For each pair, the annotation is considered reliable if at least four annotators agree on its correctness or incorrectness; cases with weaker agreement are treated as ambiguous and removed or corrected according to the refinement protocol.

The refinement statistics are presented in \cref{tab:3}. Overall, 90.6\% of the automatic annotations are verified as accurate, including 89.7\% for correct pairs and 91.5\% for incorrect pairs, demonstrating an initial high reliability brought by the INSETS pipeline. After retaining verified samples, rectifying erroneous labels, and discarding ambiguous cases, MVEI contains 3,086 image--statement pairs. As summarized in \cref{tab:4}, MVEI covers 424 distinct emotion labels, with an average of 5.2 emotion labels per image and an average statement length of 36.5 words. Compared with fully manual construction, this model-human collaborative procedure strikes a desirable balance between annotation reliability and cost, requiring 100 person-hours for the final refinement stage.

We further characterize MVEI in \cref{fig:4}. The upper flow presents a representative sample, illustrating how a raw image is enriched with multiple open-vocabulary emotion labels and then converted into an ESJ statement. The lower part of \cref{fig:4}(b) shows that MVEI covers a broad spectrum of emotion labels, with frequent labels such as \textit{Joy}, \textit{Excitement}, \textit{Nostalgia}, \textit{Happiness}, and \textit{Contentment}. When mapped to the primary emotions in POM, \textit{Joy} accounts for the largest proportion, followed by \textit{Sadness}, \textit{Love}, \textit{Fear}, \textit{Surprise}, and \textit{Anger}. The lower part of \cref{fig:4}(c) presents the distribution of constructed statements, showing balanced correct/incorrect labels, coverage of all four ESJ dimensions, and an appropriate statement-length distribution.

\section{The EmObserver Model}

\begin{figure*}[t]
    \centering
    \includegraphics[width=1\linewidth]{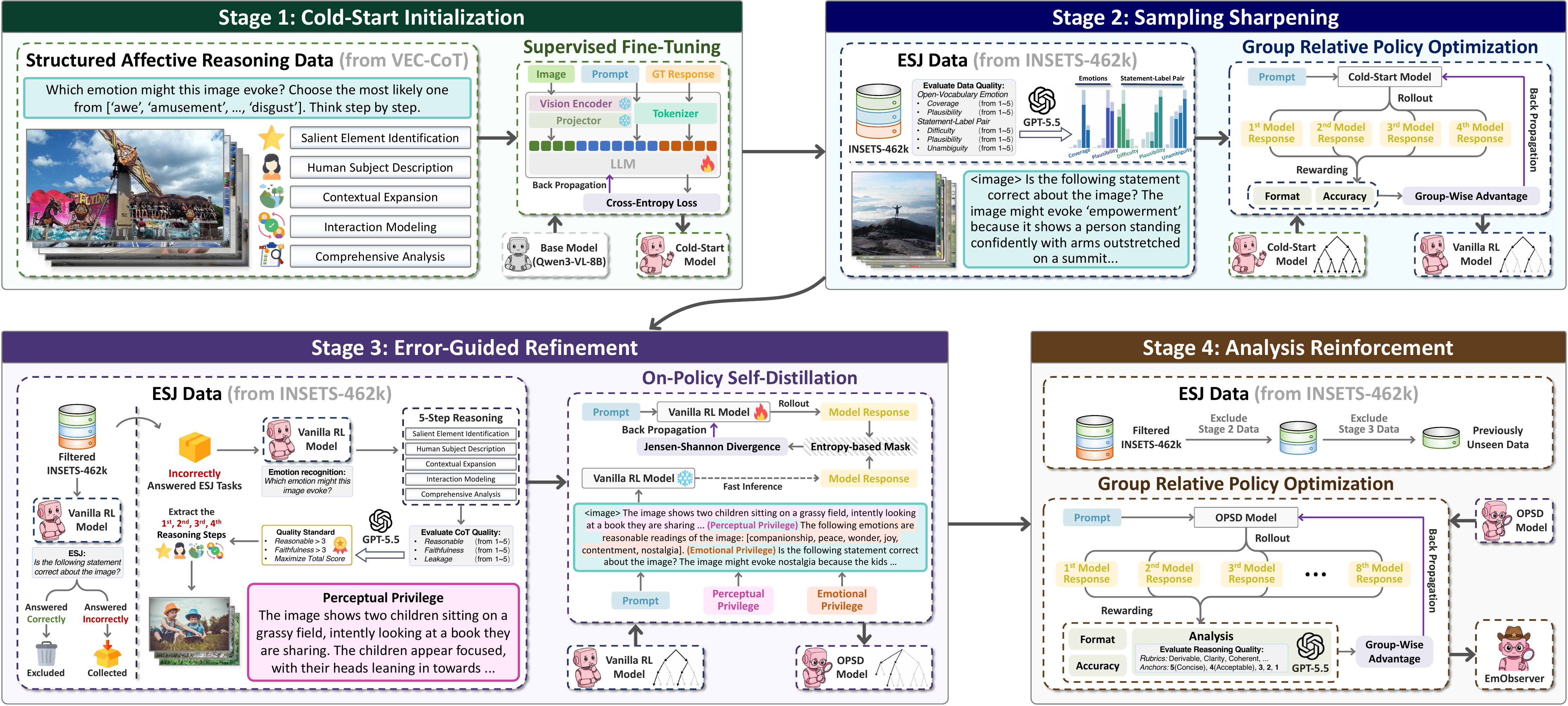}
    \vskip -0.10in
    \caption{\textbf{Overview of the EmObserver training pipeline.} Stage 1 performs cold-start supervised fine-tuning on structured affective reasoning data to initialize basic emotion reasoning ability. Stage 2 uses ESJ data from INSETS-462k and Group Relative Policy Optimization \cite{shao2024deepseekmath} to sharpen the sampling space. Stage 3 collects incorrectly answered ESJ samples and performs error-guided On-Policy Self-Distillation \cite{opsd}. Stage 4 further reinforces analysis quality on previously unseen ESJ data, yielding the final EmObserver model.}
    \label{fig:5}
    \vskip -0.10in
\end{figure*}

The ESJ formulation provides not only an ambiguity-reduced evaluation paradigm, but also an in-depth and flexible supervision format for improving visual emotional intelligence. To this end, we introduce \textbf{EmObserver}, an emotion-oriented MLLM optimized to perform diverse AICA tasks with accurate and generalizable reasoning. Its overall training pipeline is illustrated in \cref{fig:5}, with details elaborated below.

\textit{\textbf{Stage 1: Cold-Start Initialization.}}
Our training starts from a pre-trained base model, whose architecture is not altered throughout the process. Specifically, we adopt Qwen3-VL-8B-Thinking \cite{qwen3vl}, denoted as $\pi_{\theta}$. Since the INSETS pipeline does not naturally provide Chain-of-Thought (CoT) rationales, we first borrow high-quality structured CoT data from VEC-CoT \cite{emocaliber} to initialize the model's affective reasoning ability. Inspired by human cognition, VEC-CoT decomposes the reasoning process into five progressive steps: identifying salient visual elements, describing human subjects when present, focusing on contextual information, modeling interactions among visual and contextual cues, and deriving the final emotional conclusion. 

Let the dataset be $\mathcal{D}_{\mathrm{S1}}=\{(x_i,y_i^*)\}_{i=1}^{N}$, which comprises $N$ input-response pairs $(x_i,y_i^*)$. Denoting an $n$-length response sequence as: $y_i^*=(y_{i,1}^*,\ldots,y_{i,n}^*)$, the supervised fine-tuning objective of this stage is:
\begin{equation}
\mathcal{L}_{\mathrm{S1}}(\theta)
=
\mathbb{E}_{\mathcal{D}_{\mathrm{S1}}}\Big [\frac{1}{n}\sum_{t=1}^{n}
\log \pi_{\theta}\!\left(y_{i,t}^* \mid x_i,y_{i,<t}^*\right)\Big ].
\end{equation}
This yields a cold-start model equipped with a reasonable reasoning pattern transferable to subsequent optimization.

\textit{\textbf{Stage 2: Sampling Sharpening.}}
For the remaining stages, we keep optimizing the same model $\pi_{\theta}$, using INSETS-462k as the sole training source. We first remove all samples that overlap with MVEI to avoid evaluation leakage. Since reinforcement learning is computationally intensive and the remaining corpus is sufficiently large, we further perform a lightweight LLM-as-judge filtering step with GPT-5.5 \cite{gpt5} to discard potentially low-quality tasks. This preprocessing retains approximately 302k ESJ tasks, denoted as $\mathcal{D}_{\mathrm{ESJ}}$. For simplicity, we denote the ESJ input-response pair still as $(x_i,y_i^*)$.

Inspired by the success of Group Relative Policy Optimization (GRPO) \cite{shao2024deepseekmath} in affective modeling \cite{emotionreasoner}, we adopt it here to sharpen the sampling behavior by incentivizing deliberation among plausible answers. We randomly sample one-tenth of $\mathcal{D}_{\mathrm{ESJ}}$ for this stage, denoted as $\mathcal{D}_{\mathrm{S2}}$. For each input $x_i\in\mathcal{D}_{\mathrm{S2}}$, the model $\pi_{\theta}$ rollouts a group of $G$ responses $\{o_g\}_{g=1}^{G}$. Each response receives two basic rewards: a format reward $\mathcal{R}_{\mathrm{format}}$ of 0.5 if it follows the required output format, and an additional accuracy reward $\mathcal{R}_{\mathrm{acc}}$ of 0.5 if the final judgement is correct. The total reward is thereby
$
\mathcal{R}(o_g)=\mathcal{R}_{\mathrm{format}}(o_g)+\mathcal{R}_{\mathrm{acc}}(o_g).
$

The group-wise advantage of response $o_g$ is subsequently computed as:
$A_g=(\mathcal{R}(o_g)-\text{mean}(\mathcal{R}))/\text{std}(\mathcal{R}).$
To balance training efficiency and on-policy exploration, responses are generated by an old policy $\pi_{\text{old}}$. Finally, this stage updates the model with a PPO-style clipped surrogate objective \cite{ppo}:
\begin{equation}
\label{eq:2}
\begin{aligned}
&\mathcal{L}_{\mathrm{S2}}(\theta)
=
\mathbb{E}_{\mathcal{D}_{\mathrm{S2}}}
\bigg[
\mathbb{E}_{o_g \sim \pi_{\mathrm{old}}}
\Big[
\mathbb{E}_{g}
\bigg[
\frac{1}{|o_g|}
\sum_{t=1}^{|o_g|}
\Big(
\min\!\bigl(
\rho_{g,t}(\theta)A_g,
\\
&\qquad
\mathrm{clip}\!\left(\rho_{g,t}(\theta),1\pm\epsilon\right)A_g
\bigr)
-\beta\,\mathbb{D}_{KL}\!\left(\pi_{\theta}\,\|\,\pi_{\mathrm{ref}}\right)
\Big)
\bigg]
\Big]
\bigg].
\end{aligned}
\end{equation}
$\rho_{g,t}(\theta)=\pi_{\theta}(o_{g,t}\mid x_i,o_{g,<t})/\pi_{\mathrm{old}}(o_{g,t}\mid x_i,o_{g,<t})$ denotes the token-level importance ratio, which corrects the discrepancy between the current and the old policy. The clipping operation stabilizes policy updates, while the KL regularization term keeps $\pi_{\theta}$ close to the reference policy $\pi_{\mathrm{ref}}$. This stage produces a vanilla RL model that samples more efficiently in AICA.

\textit{\textbf{Stage 3: Error-Guided Refinement.}}
Despite its direct effectiveness, GRPO is considered constrained by the latent response space induced by the cold-start model under each given prompt \cite{does}. This drives us to explore On-Policy Self-Distillation (OPSD) \cite{opsd}, where a teacher branch is provided with privileged information and its input-induced improvement is distilled back into the same policy. Formally, given an input $x_i$ and privileged information $p_i$, the student branch samples a response from the original policy: $o_i\sim\pi_{\theta}(\cdot\mid x_i)$. The student distribution is computed as $\sigma_{i,t}^{S}=\pi_{\theta}(\cdot\mid x_i,o_{i,<t})$, while the teacher distribution is evaluated on the same tokens under the privileged condition: $\sigma_{i,t}^{T}=sg[\pi_{\theta}(\cdot\mid x_i,p_i,o_{i,<t})]$, where $sg[\cdot]$ denots the stop-gradient operation. The student distribution is then encouraged to approach the teacher distribution through a KL-type target.

The keys to effective OPSD lie in suitable privileged information and informative data. Specifically, we first run the vanilla RL model on $\mathcal{D}_{\mathrm{ESJ}}$ and collect the incorrectly answered cases:
$
\mathcal{D}_{\mathrm{fail}}
=
\{(x_i,y_i^*)\in\mathcal{D}_{\mathrm{ESJ}}
\mid \hat{y}_i \neq y_i^*\},    
$
where $\hat{y}_i$ denotes the model prediction. For these failed cases, we further require the model to infer the image-evoked emotion without restricting the answer space. By adopting a formulation closer to the cold-start target, this step encourages the model to expose latent cues that were underutilized in the original ESJ decision, thereby providing informative signals for subsequent distillation.

After obtaining these responses, we exploit their structured reasoning format to remove the final step that summarizes the emotion decision. The remaining reasoning components are less coupled with specific tasks, yet preserve rich visual perception and contextual analysis. After filtering them with a lightweight LLM-as-judge process using GPT-5.5, the retained high-quality cases are used as \textit{perceptual privilege}, denoted as $p_i^{\mathrm{perc}}$. In parallel, we introduce the open-vocabulary emotions obtained in the first stage of INSETS as complementary \textit{emotional privilege}, denoted as $p_i^{\mathrm{emo}}$. The combined privilege for input $x_i$ is then defined as
$
p_i=\{p_i^{\mathrm{perc}},p_i^{\mathrm{emo}}\}.
$
Consequently, we further select training samples for which the teacher branch corrects the previous error with prediction $\hat{y}_i^{t}$:
\begin{equation}
\mathcal{D}_{\mathrm{S3}}
=
\{(x_i,y_i^*,p_i)\mid (x_i,y_i^*)\in\mathcal{D}_{\mathrm{fail}},
\ \hat{y}_i^{t}=y_i^*\}.
\end{equation}

To focus optimization on highly reliable signals, we additionally apply an entropy-based token mask. Let $H_{i,t}^{T}$ denote the entropy of the teacher distribution at token position $t$. We only compute the distillation loss on positions where $H_{i,t}^{T}$ falls within the lower half of the sequence-level entropy distribution:
$
m_{i,t}=
\mathbb{I}\!\left[
H_{i,t}^{T}\leq \mathrm{median}\left(\{H_{i,k}^{T}\}_{k=1}^{|o_i|}\right)
\right],
$
where $\mathbb{I}[\cdot]$ is the indicator function.  Finally, adopting the Jensen-Shannon divergence $\mathbb{D}_{JS}$, we acquire the OPSD model through the following training objective:
\begin{equation}
\mathcal{L}_{\mathrm{S3}}(\theta)
=
\mathbb{E}_{\mathcal{D}_{\mathrm{S3}}}
\bigg [
\mathbb{E}_{o_i\sim\pi_{\theta}(\cdot\mid x_i)}
\Big [
\mathbb{E}_t
\big [
m_{i,t}\mathbb{D}_{JS}\!\left(
\sigma_{i,t}^{S}
\,\|\, 
\sigma_{i,t}^{T}
\right)
\big]
\Big]
\bigg].
\end{equation}

\newcommand{\szg}[1]{{\color{gray}\scriptsize #1}}   
\newcommand{\Lbar}[1]{\multicolumn{6}{c}{\cellcolor[HTML]{ebebeb}#1}}  
\newcommand{\Rbar}[1]{\multicolumn{6}{c}{\cellcolor[HTML]{ebebeb}#1}}            
\newcommand{\blank}{\multicolumn{6}{c}{}}                              
\definecolor{oursblue}{HTML}{e1e2ff}
\begin{table*}[t]
\caption{Performance of advanced MLLMs across scales and purposes on MVEI. \textbf{Bold} and \underline{underline} denote the best and second-best results within each MLLM category, respectively.}
\label{tab:5}
\vskip -0.1in
\centering
\resizebox{\textwidth}{!}{
\renewcommand{\arraystretch}{1.1}
\begin{tabular}{lccccclccccc}
\toprule
\multirow{2}{*}{\bf Models} & \multicolumn{5}{c}{MVEI: Accuracy (\%)} &
\multirow{2}{*}{\bf Models} & \multicolumn{5}{c}{MVEI: Accuracy (\%)} \\
\cmidrule(lr){2-6} \cmidrule(lr){8-12}
 & \makebox[0.04\linewidth][c]{SP} & \makebox[0.04\linewidth][c]{EI} & \makebox[0.04\linewidth][c]{SC} & \makebox[0.04\linewidth][c]{PS} & \makebox[0.04\linewidth][c]{Total}
 & & \makebox[0.04\linewidth][c]{SP} & \makebox[0.04\linewidth][c]{EI} & \makebox[0.04\linewidth][c]{SC} & \makebox[0.04\linewidth][c]{PS} & \makebox[0.04\linewidth][c]{Total} \\
\midrule
\Lbar{Small-Size (7--9B) Open-Source MLLMs}            & Qwen3-VL-Thinking \cite{qwen3vl} \szg{32B}      & 84.21 & 81.20 & 76.80 & 74.71 & 78.87 \\
DeepSeek-VL \cite{arxiv2024deepseekvl} \szg{7B}          & 77.19 & 68.06 & 79.95 & 71.10 & 72.94 & Qwen3.5 \cite{qwen3.5} \szg{27B}             & 82.96 & 77.28 & 75.49 & 75.87 & 77.22 \\
mPLUG-Owl3 \cite{arxiv2024mplugowl3} \szg{7B}          & 74.19 & 79.98 & 71.82 & 60.77 & 72.39 & Gemma-4 \cite{gemma4} \szg{31B}             & \underline{86.47} & \textbf{81.64} & \textbf{82.57} &	\underline{80.39} &	\textbf{82.18} \\
MiniCPM-V-2.6 \szg{8B} \cite{arxiv2024minicpm}      & 75.94 & 75.11 & 77.85 & 59.23 & 71.91 & Qwen3.6 \cite{qwen3.6-27b} \szg{27B}             & 77.69 & 73.11 & 72.74 & 67.35 & 72.16 \\
Qwen2.5-VL-Instruct \cite{2025qwen25vl} \szg{7B} & 73.43 & 80.68 & 73.39 & 60.77 & 72.94 & \Rbar{Large-Size (72--78B) Open-Source MLLMs} \\
InternVL3 \cite{internvl3} \szg{8B}          & 75.44 & 77.37 & 81.13 & 73.16 & 76.99 & Qwen2.5-VL-Instruct \cite{2025qwen25vl} \szg{72B} & \underline{79.20} & \underline{82.51} & \underline{77.33} & \textbf{80.39} & \underline{80.27} \\
GLM-4.1V-Thinking \cite{glm41v} \szg{9B}  & \textbf{83.46} & 81.03 & \underline{81.65} & \underline{73.42} & \underline{79.59} & InternVL3 \cite{internvl3} \szg{78B}           & \textbf{83.46} & \textbf{82.94} & \textbf{79.69} & \underline{79.61} & \textbf{81.37} \\
MiMo-VL-RL \cite{mimovl} \szg{7B}         & 65.16 & 78.94 & 78.64 & 69.55 & 74.72 & \Rbar{Proprietary MLLMs} \\
InternVL-3.5 \cite{internvl35} \szg{8B}       & 74.19 & 78.94 & 78.77 & 64.65 & 74.69 & Kimi-K2.5 \cite{kimik25}   & 79.95 & \underline{80.33} & 79.03 & \underline{81.55} & \underline{80.27} \\
MiniCPM-V-4.5 \cite{minicpmv45} \szg{8B}      & 76.94 & \underline{81.11} & 75.62 & 64.65 & 75.08 & Seed-2.0-Pro \cite{seed20pro}                 & \textbf{87.22} & \textbf{82.25} & \underline{79.16} & \textbf{82.58} & \textbf{82.21} \\
Qwen3-VL-Instruct \szg{8B} \cite{qwen3vl} & 74.69 & \underline{81.11} & \textbf{81.78} & 60.39 & 75.24 & Qwen3.6-Plus \cite{qwen3.6-plus}                 & \underline{85.46} & 76.24 & 74.97 & 78.45 & 77.67 \\
Qwen3-VL-Thinking \cite{qwen3vl} \szg{8B}     & \underline{83.21} & 76.41 & 79.16 & 64.13 & 74.89 & GPT-5.5 \cite{gpt5}                      & 81.70 & 75.54 & \textbf{81.65} & 79.48 & 78.84 \\
GLM-4.6V-Flash \cite{glm41v} \szg{9B}     & 81.45 & \textbf{81.90} & 80.21 & \textbf{76.65} & \textbf{80.10} & \Rbar{Emotion-Oriented MLLMs} \\
Qwen3.5 \cite{qwen3.5} \szg{9B}            & 76.44 & 75.28 & 64.61 & 66.19 & 70.51 & EmoVIT \cite{cvpr2024emovit} \szg{7B}               & 67.17 & \underline{79.20} & 74.44 & 56.65 & 70.80 \\
LLaVA-OneVision-2 \cite{llava-onevision2} \szg{8B}  & 79.45 & 77.28 & 74.71 & 65.42 & 73.95 & Emotion-Qwen \cite{emotion-qwen} \szg{7B}             & \underline{73.93} & 75.46 & 70.51 & \underline{67.87} & \underline{72.13} \\
\Lbar{Middle-Size (25--38B) Open-Source MLLMs}        & EmoCaliber \cite{emocaliber} \szg{7B}           & 69.92 & 68.06 & \underline{79.69} & 66.84 & 70.87 \\
Qwen2.5-VL-Instruct \cite{2025qwen25vl} \szg{32B} & 82.96 & 80.07 & 76.80 & 73.94 & 78.09 & \cellcolor{oursblue}EmObserver (\textbf{Ours}) \szg{8B} & \cellcolor{oursblue}\textbf{86.47} & \cellcolor{oursblue}\textbf{83.55} & \cellcolor{oursblue}\textbf{90.30} & \cellcolor{oursblue}\textbf{86.06} & \cellcolor{oursblue}\textbf{86.23} \\
Gemma-3 \cite{gemma3} \szg{27B}           & 82.96 & 78.33 & 74.57 & 73.94 & 77.45 & \Rbar{Human Baselines} \\
InternVL3 \cite{internvl3} \szg{38B}         & \textbf{89.72} & \underline{81.46} & 79.29 & \textbf{81.16} & \underline{81.92} & Human  \szg{Avg}            & 92.32 & 90.11 & 95.27 & 89.60 & 91.64 \\
Qwen3-VL-Instruct \cite{qwen3vl} \szg{32B}     & 81.95 & 81.38 & \underline{80.34} & 65.42 & 77.19 & Human  \szg{Best}           & 97.37 & 95.95 & 98.67 & 94.69 & -- \\
\bottomrule
\end{tabular}}
\vskip -0.10in
\end{table*}

\begin{figure}[t]
    \centering
    \includegraphics[width=1\linewidth]{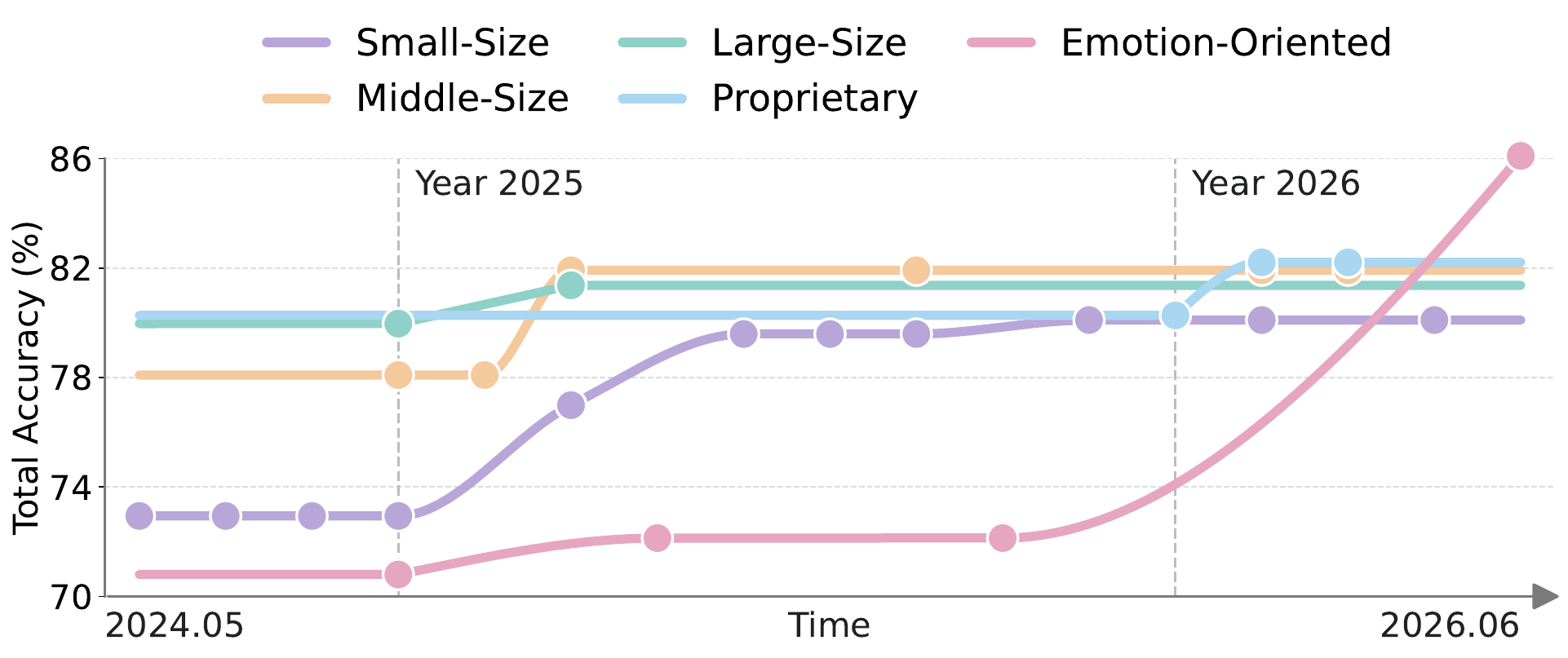}
    \vskip -0.10in
    \caption{Evolution of the best-performing model in each MLLM category on MVEI from May 2024 to June 2026. Each curve traces the running-best Total accuracy attained within a category up to each time point.}
    \label{fig:6}
    \vskip -0.10in
\end{figure}

\textit{\textbf{Stage 4: Analysis Reinforcement.}}
Through preliminary evaluation, we observe that the OPSD model improves overall ESJ performance but also exhibits several undesirable behaviors. A salient issue is the occasional inconsistency between the reasoning process and the final answer. We hypothesize that this occurs because the privileged teacher signal may sometimes provide sharper optimization pressure on answer tokens than on the reasoning path, gradually inducing a small number of misaligned cases where the answer is correct but the analysis does not faithfully support it. To mitigate this issue, we apply GRPO similar to Stage 2 on an additional ESJ subset that the model has not seen:
$
\mathcal{D}_{\mathrm{S4}}
\subset
\mathcal{D}_{\mathrm{ESJ}}
\setminus
\left(\mathcal{D}_{\mathrm{S2}}\cup\mathcal{D}_{\mathrm{S3}}\right).
$
We introduce an LLM-as-judge consistency reward to encourage both the analysis itself and analysis-answer alignment with the core prompt shown below:

\begin{tcolorbox}[
    colback=PromptBlue,
    colframe=PromptBorder,
    boxrule=0.7pt,
    arc=2pt,
    left=2pt,
    right=2pt,
    top=2pt,
    bottom=2pt,
    title=\textbf{LLM-as-Judge Prompt},
    coltitle=black,
    colbacktitle=PromptGray,
    fonttitle=\small\bfseries
]
\footnotesize
You are a strict reasoning-quality judge for a multimodal emotion-statement verification task.
The student model is shown an image (or several images) and a natural-language STATEMENT that asserts something about the emotional content of the image(s).
The student must decide whether the statement is correct.

Definitions used below:
\begin{itemize}
    \item \textbf{derivable}: a careful reader could reach the \texttt{<answer>} verdict by only following the \texttt{<analysis>} content; no logical jump outside the analysis is required.
    \item \textbf{clear path}: the analysis explicitly addresses whether the STATEMENT is correct, names the key evidence, and reaches an explicit verdict; it is not merely emotion description.
    \item \textbf{coherent}: sentences flow logically, with no self-contradiction inside the analysis and no irrelevant tangents.
    \item \textbf{answer mapping}: \texttt{A} or \texttt{correct} means the model claims the STATEMENT is correct; \texttt{B} or \texttt{incorrect} means the model claims the STATEMENT is incorrect; other phrasings should be inferred from natural language.
\end{itemize}
\end{tcolorbox}
Following the previous notation, for each rollout $o_g$, we augment the reward with a normalized GPT-5.5-derived consistency term:
$
\mathcal{R}(o_g)
=
\mathcal{R}_{\mathrm{format}}(o_g)
+
\mathcal{R}_{\mathrm{acc}}(o_g)
+
\mathcal{R}_{\mathrm{cons}}(o_g).
$
All other optimization components remain unchanged. Optimizing the OPSD model on $\mathcal{D}_{\mathrm{S4}}$ with the same objective as in \cref{eq:2} yields the final EmObserver model.

\textit{\textbf{Implementation Details.}}
\textit{In Stage 1}, $\mathcal{D}_{\mathrm{S1}}$ contains 143,446 samples. We train for 1 epoch with a batch size of 32 and a learning rate of $1\times10^{-5}$, while freezing the vision encoder.
\textit{In Stage 2}, $\mathcal{D}_{\mathrm{S2}}$ contains 30,191 samples. We use a batch size of $8$ with gradient accumulation of $2$, sample $4$ responses for each prompt, and reuse each generation batch for two optimization steps. The KL coefficient is set to $\beta=0.04$, the sampling temperature is $1.0$, and the model is trained for 1 epoch with a learning rate of $1\times10^{-6}$.
\textit{In Stage 3}, $\mathcal{D}_{\mathrm{S3}}$ contains 4,502 samples. We use a batch size of $8$ with gradient accumulation of $8$, set the temperature to $1.0$, and compute the distillation loss over the top-128 logits at each token position. The model is trained for 3 epochs with a learning rate of $2\times10^{-6}$, and the vision encoder is frozen in this stage.
\textit{In Stage 4}, $\mathcal{D}_{\mathrm{S4}}$ contains 5,502 samples. We sample $8$ responses for each prompt and do not reuse generations across optimization steps. All other hyperparameters are identical to Stage 2.

\begin{figure*}[t]
    \centering
    \includegraphics[width=1\linewidth]{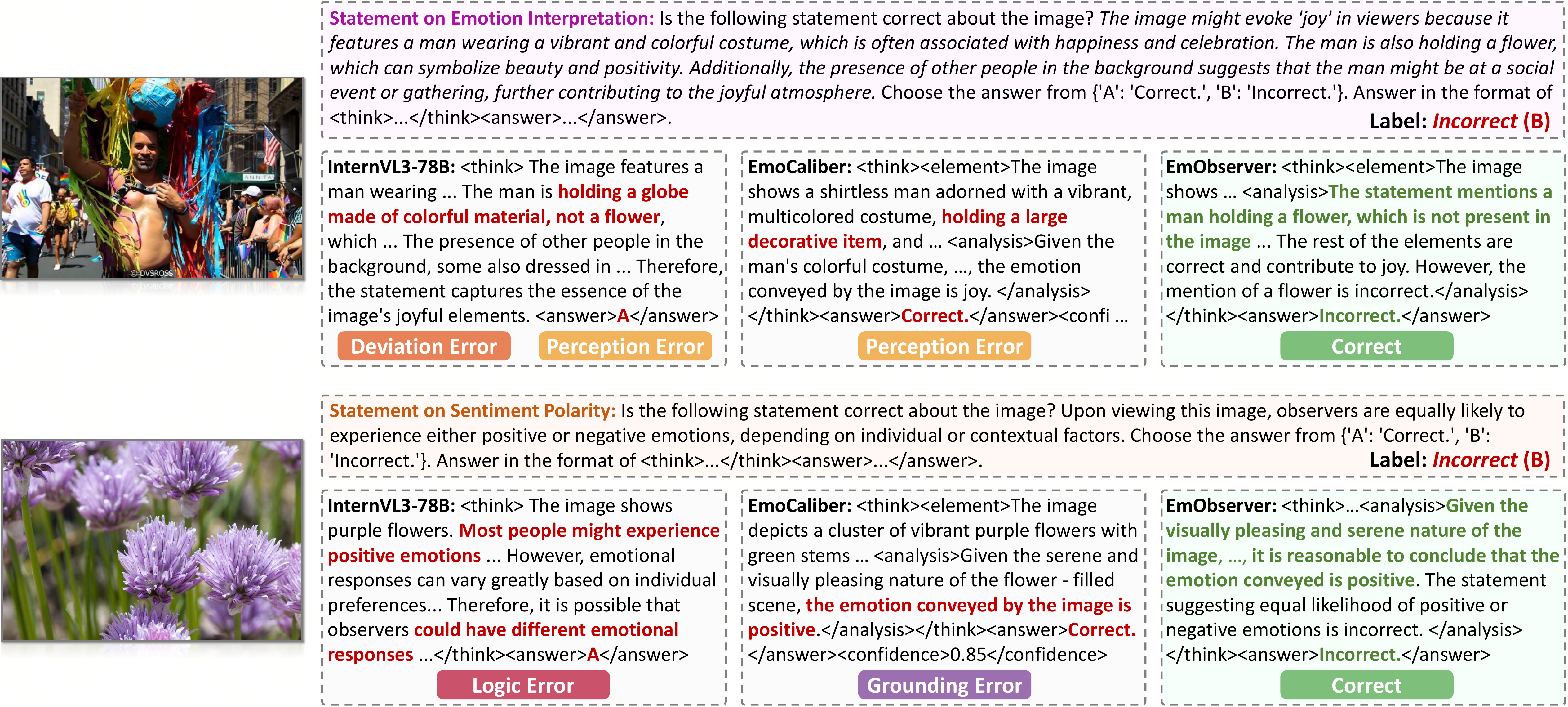}
    \vskip -0.10in
    \caption{Qualitative comparison of representative MLLMs on two MVEI cases. Each row shows a statement, and the responses of InternVL3-78B \cite{internvl3}, EmoCaliber \cite{emocaliber}, and EmObserver, annotated with the corresponding error type. \textcolor{red}{Red}/\textcolor{green!50!black}{green} highlight flawed/correct reasoning. Unlike the baselines, EmObserver performs explicit analysis of the statement and aligns its verdict with that analysis, avoiding perception, deviation, logic, and grounding errors.}
    \label{fig:7}
    \vskip -0.10in
\end{figure*}

\begin{figure}[t]
    \centering
    \includegraphics[width=1\linewidth]{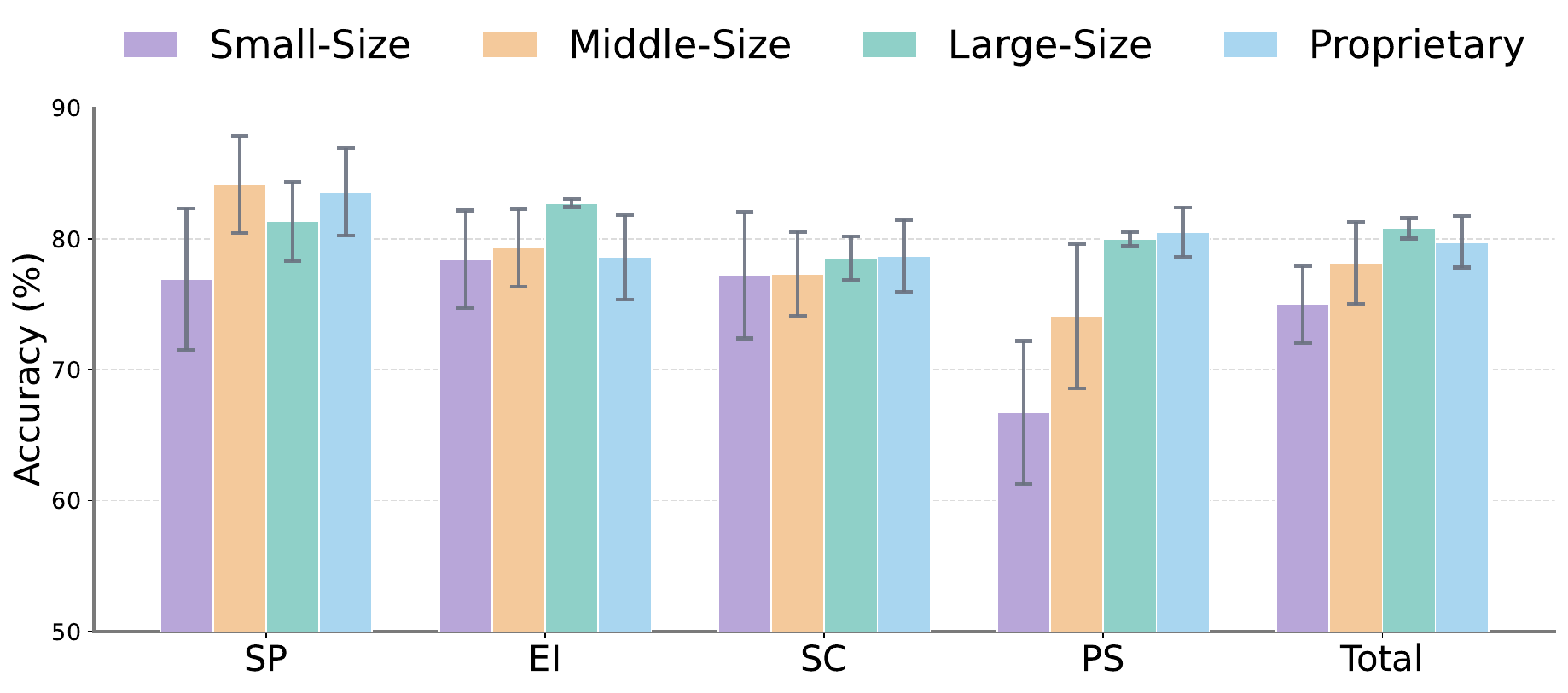}
    \vskip -0.10in
    \caption{Per-dimension accuracy of four general-purpose MLLM categories on MVEI. Each bar reports the mean over models within a category, and error bars indicate the standard deviation across them.}
    \label{fig:8}
    \vskip -0.10in
\end{figure}

\section{Experiments}
This section presents a comprehensive empirical study of visual emotional intelligence in MLLMs, covering both quantitative comparisons and qualitative analyses. The evaluation extends beyond the proposed ESJ task and MVEI benchmark to related AICA datasets, covering diverse scenarios.

\subsection{Evaluation on MVEI}
\textit{\textbf{Overall Comparison (\cref{tab:5}).}}
MVEI evaluates visual emotional intelligence along four dimensions: sentiment polarity (SP), emotion interpretation (EI), scene context (SC), and perception subjectivity (PS), and we use these abbreviations throughout this section. All MLLMs are queried with the exact prompt in \cref{section:3}. We also conduct a human study with 25 participants on a 300-sample subset, reporting the average participant performance as \textit{Human Avg} and the best participant score on each dimension as \textit{Human Best}. The results reveal a highly uneven capability profile among existing MLLMs: different models excel on different dimensions, and none of the prior models achieves consistently dominant performance across all categories. This suggests that visual emotional intelligence is not a monolithic ability, but rather requires a coordinated set of affective, contextual, and subjectivity-aware reasoning skills. Against this backdrop, EmObserver shows the strongest and most balanced performance, achieving \textbf{86.23\%} total accuracy and obtaining the best model-side results on three dimensions. Nevertheless, the remaining gap to human performance indicates significant room for improvement.

\textit{\textbf{Temporal Evolution of MLLMs (\cref{fig:6}).}}
Beyond the static comparison, we further trace the evolution of the best performance within each model category. Over the last two years, open-source MLLMs have improved steadily across scales, whereas early emotion-oriented MLLMs have lagged behind. Meanwhile, the gaps among different model scales have substantially narrowed, possibly benefiting from the growing practice of distillation across models and scales. Nevertheless, general-purpose MLLMs have begun to show signs of saturation, suggesting that model recency alone may not reliably translate into stronger visual emotional intelligence.

\textit{\textbf{Per-Dimension Comparison (\cref{fig:8}).}}
To further shed light on dimension-wise variations, we compare MLLM categories along each ESJ dimension. The results show that category-level advantages vary substantially across dimensions: middle-size MLLMs perform strongly on SP, large-size MLLMs lead on EI, and the gap on SC is relatively modest. The most pronounced separation appears on PS, where small-size MLLMs lag clearly behind larger and proprietary models, suggesting that perception subjectivity may demand richer world knowledge and social-emotional reasoning. In terms of total accuracy, performance generally improves from small to middle and large open-source models, while proprietary models do not consistently surpass the strongest open-source category. This again indicates that emotional intelligence remains underemphasized in proprietary systems, given their otherwise leading capabilities in general vision-language tasks \cite{gpt5,seed20pro}.

\newcommand{\dlt}[1]{\,{\color{gray}\scriptsize $-$#1}}   

\begin{table*}[t]
\caption{Performance of MLLMs on MVEI under perturbed input images. Gray subscripts denote the accuracy drop relative to the original (unperturbed) setting. \textit{Positive Ratio} is the proportion of responses judged ``Correct'', while \textit{Give-up Ratio} is the proportion of cases in which the model fails to commit to either judgement.}
\label{tab:6}
\vskip -0.1in
\centering
\resizebox{0.82\textwidth}{!}{
\renewcommand{\arraystretch}{1.2}
\begin{tabular}{lccccccc}
\toprule
\multirow{2}{*}{\bf Models} & \multicolumn{5}{c}{MVEI: Accuracy (\%)} &
\multirow{2}{*}{\makecell{Positive\\Ratio}} & \multirow{2}{*}{\makecell{Give-up\\Ratio}} \\
\cmidrule(lr){2-6}
 & \makebox[0.08\linewidth][c]{SP} & \makebox[0.08\linewidth][c]{EI} & \makebox[0.08\linewidth][c]{SC} & \makebox[0.08\linewidth][c]{PS} & \makebox[0.08\linewidth][c]{Avg} & & \\
\midrule
Random Guess & 50.00 & 50.00 & 50.00 & 50.00 & 50.00 & 50.00 & 0 \\
Qwen3-VL-Thinking \cite{qwen3vl} \szg{8B} & 47.12\dlt{36.09} & 44.56\dlt{31.85} & 49.02\dlt{30.14} & 47.35\dlt{16.78} & 46.69\dlt{28.20} & 4.15 & 0 \\
GLM-4.6V-Flash \cite{glm41v} \szg{9B} & 39.60\dlt{41.85} & 45.26\dlt{36.64} & 69.46\dlt{10.75} & 50.97\dlt{25.68} & 51.94\dlt{28.16} & 40.57 & 4.96 \\
Gemma-4 \cite{gemma4} \szg{31B} & 42.61\dlt{43.86} & 44.82\dlt{36.82} & 47.71\dlt{34.86} & 47.61\dlt{32.78} & 45.95\dlt{36.23} & 3.99 & 0 \\
InternVL3 \cite{internvl3} \szg{78B} & 39.60\dlt{43.86} & 44.13\dlt{38.81} & 56.62\dlt{23.07} & 58.06\dlt{21.55} & 50.13\dlt{31.24} & 26.70 & 0 \\
EmoCaliber \cite{emocaliber} \szg{7B} & 33.33\dlt{36.59} & 55.70\dlt{12.36} & 68.02\dlt{11.67} & 40.77\dlt{26.07} & 52.11\dlt{18.76} & 62.70 & 5.18 \\
GPT-5.5 \cite{gpt5} & 48.87\dlt{32.83} & 44.56\dlt{30.98} & 52.23\dlt{29.42} & 49.68\dlt{29.80} & 48.87\dlt{29.97} & 5.51 & 0 \\
\cellcolor{oursblue}EmObserver (\textbf{Ours}) \szg{8B} & \cellcolor{oursblue}40.85\dlt{45.62} & \cellcolor{oursblue}44.82\dlt{38.73} & \cellcolor{oursblue}67.53\dlt{22.77} & \cellcolor{oursblue}56.90\dlt{29.16} & \cellcolor{oursblue}52.53\dlt{33.70} & \cellcolor{oursblue}27.25 & \cellcolor{oursblue}0 \\
\bottomrule
\end{tabular}}
\vskip -0.16in
\end{table*}

\begin{figure}[t]
    \centering
    \includegraphics[width=1\linewidth]{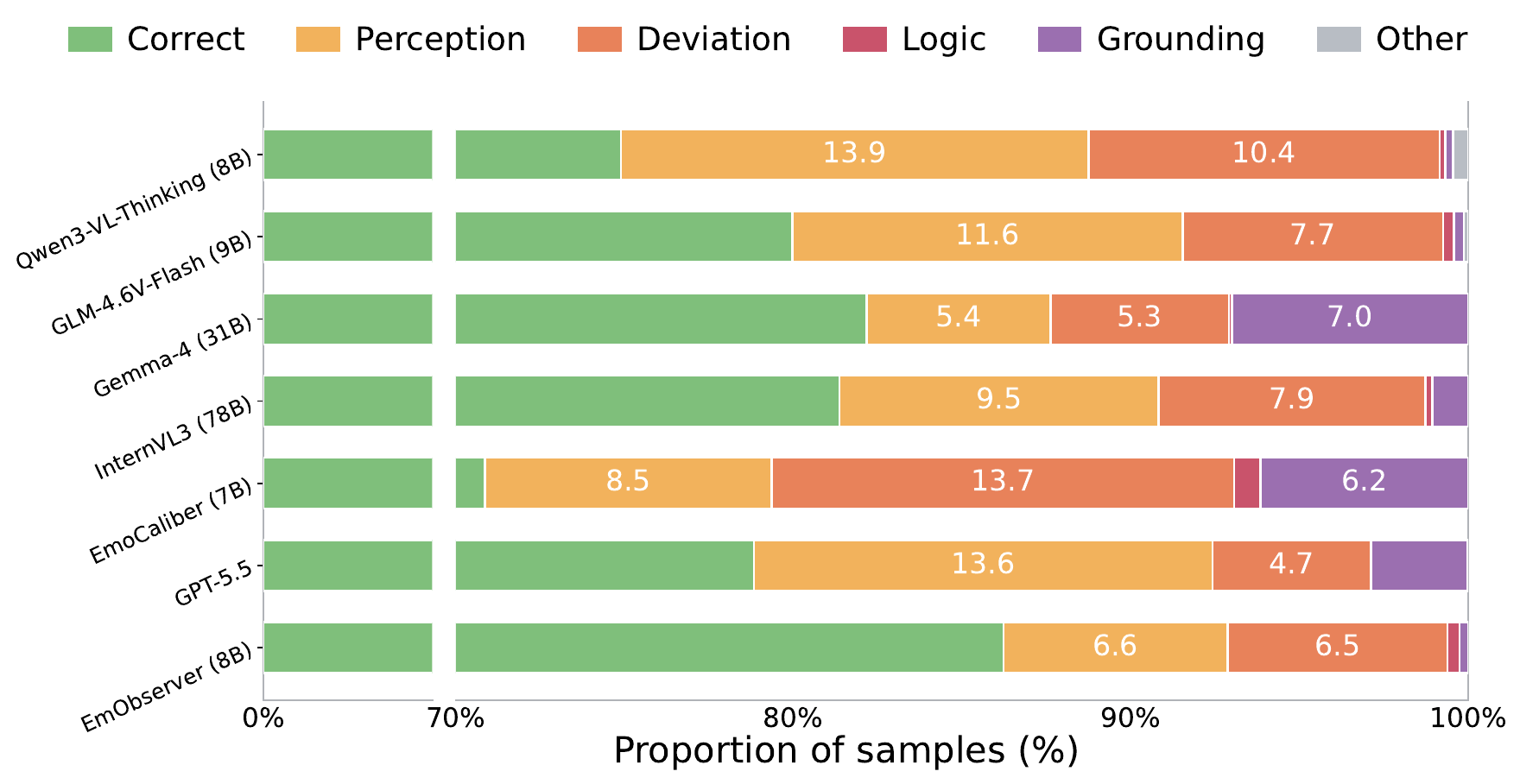}
    \vskip -0.10in
    \caption{Distribution of error types across MLLMs on MVEI. Each bar shows the proportion of correct predictions and the five error categories. X-axis is truncated at 70\% to highlight the error regions.}
    \label{fig:9}
    \vskip -0.16in
\end{figure}

\textit{\textbf{Error-Oriented Analysis (\cref{fig:7} \& \cref{fig:9}).}}
We conduct further qualitative and quantitative error analyses to understand why MLLMs fail. We categorize incorrect predictions into four major types. \textit{Perception Error} refers to cases where reasoning misinterprets visual evidence. \textit{Logic Error} denotes internally inconsistent or self-contradictory reasoning. \textit{Deviation Error} occurs when the model draws an incorrect intermediate conclusion and is subsequently led along a wrong path. \textit{Grounding Error} indicates that the final answer is not faithfully grounded in the preceding reasoning. \cref{fig:7} illustrates how advanced MLLMs exhibit these errors on ESJ tasks. Even high-performing models may produce superficially plausible but flawed reasoning, leading to incorrect or weakly justified answers. This highlights that reliable reasoning is essential not only for accuracy, but also for the interpretability and traceability of affective decisions.

Based on this taxonomy, we further use an LLM-as-judge protocol with Kimi-K2.5 \cite{kimik25} to systematically annotate error types for representative MLLMs on MVEI, as statistically shown in \cref{fig:9}. The \textit{Other} category denotes rare errors that are difficult to fit in the taxonomy. Overall, existing models mainly fail due to perception, deviation, and grounding errors, while logic errors are relatively uncommon, reflecting the improved reasoning coherence of recent MLLMs. EmObserver shows a comprehensive reduction across error types: it nearly eliminates grounding errors and approaches among the lowest proportions of perception and deviation errors. These results suggest that its training strategy improves not only final accuracy but also the faithfulness and stability of reasoning. Meanwhile, the remaining errors highlight visual perception and early reasoning deviation as two key directions for further improving visual emotional intelligence.

\begin{figure}[t]
    \centering
    \includegraphics[width=1\linewidth]{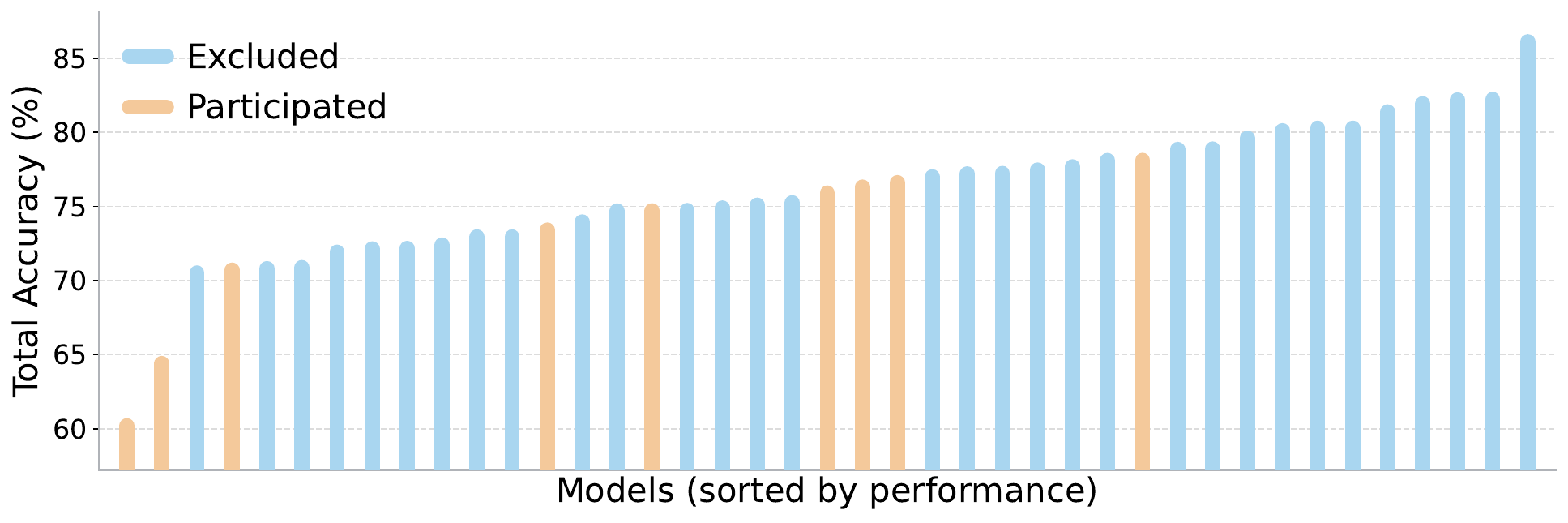}
    \vskip -0.10in
    \caption{Total accuracy of MLLMs on MVEI, sorted in ascending order. Colors indicate whether models participated in the dataset construction or not.}
    \label{fig:10}
    \vskip -0.10in
\end{figure}

\textit{\textbf{Diagnostic Evaluation (\cref{tab:6}).}}
To probe the potential influence of language priors, we conduct a diagnostic evaluation by replacing the input image with an uninformative black image while keeping the statement unchanged. As the results show, all MLLMs, including EmObserver, experience substantial drops and approach random-guessing accuracy, indicating that previous conclusions do not stem from statement-answer shortcuts. Two observations are noteworthy. First, several MLLMs still achieve relatively high SC accuracy, sometimes approaching 70\%, suggesting that scene context statements can provide textual priors. We consider this reasonable, since context is exactly intended to supply situational information that may shape emotional interpretation. Yet all MLLMs still drop by more than 10.75\% on SC after removing the image, confirming that visual evidence remains essential. Second, some MLLMs exhibit a strong bias toward negative judgements, reflected by the \textit{Positive Ratio}. This may stem from the black image itself, which can evoke emptiness or negative affective associations, a bias difficult to fully avoid. Overall, this diagnostic evaluation confirms that ESJ formulation does not introduce strong language priors that weaken the task.

\textit{\textbf{Bias of Data-Constructors (\cref{fig:10}).}}
Although most MLLMs participating in dataset construction are deliberately excluded from our main evaluation, we further visualize their relative positions among all evaluated models. It can be observed that these MLLMs do not exhibit a systematic advantage. This confirms that distributing generation across multiple MLLMs substantially mitigates construction-induced unfairness, including potential advantages that may propagate to later models within the same families.

\newcommand{\Cbar}[1]{\multicolumn{10}{c}{\cellcolor[HTML]{ebebeb}#1}}

\begin{table*}[t]
\caption{Comparison of advanced MLLMs across scales and purposes on EEmo-Bench (Perception Split) and VECBench.}
\label{tab:7}
\vskip -0.1in
\centering
\resizebox{0.94\textwidth}{!}{
\renewcommand{\arraystretch}{1.1}
\begin{tabular}{lccccccccc}
\toprule
\multirow{2}{*}{\bf Models} & \multicolumn{5}{c}{EEmo-Bench (Perception Split): Accuracy (\%)} & \multicolumn{4}{c}{VECBench: Accuracy (\%)} \\
\cmidrule(lr){2-6} \cmidrule(lr){7-10}
 & \makebox[0.055\linewidth][c]{Single-Y/N} & \makebox[0.055\linewidth][c]{Single-W/H} & \makebox[0.055\linewidth][c]{Pair-Y/N} & \makebox[0.055\linewidth][c]{Pair-W/H} & \makebox[0.055\linewidth][c]{Avg}
 & \makebox[0.055\linewidth][c]{ID-VER} & \makebox[0.055\linewidth][c]{ID-VSA} & \makebox[0.055\linewidth][c]{OOD-VER} & \makebox[0.055\linewidth][c]{Avg} \\
\midrule
\Cbar{Small-Size (7--9B) Open-Source MLLMs} \\
DeepSeek-VL \cite{arxiv2024deepseekvl} \szg{7B} & 61.89 & 48.06 & 56.96 & 35.39 & 50.57 & 39.30 & \textbf{83.15} & 43.00 & 55.15 \\
mPLUG-Owl3 \cite{arxiv2024mplugowl3} \szg{7B} & 60.92 & 49.03 & 55.37 & 47.32 & 53.16 & 37.75 & 75.90 & 33.96 & 49.20 \\
MiniCPM-V-2.6 \szg{8B} \cite{arxiv2024minicpm} & \underline{65.10} & 49.23 & 57.85 & 53.48 & 56.42 & 39.65 & 76.50 & 45.66 & 53.94 \\
Qwen2.5-VL-Instruct \cite{2025qwen25vl} \szg{7B} & 60.97 & 47.09 & 58.05 & 51.59 & 54.43 & 41.55 & 78.45 & 49.23 & 56.41 \\
InternVL3 \cite{internvl3} \szg{8B} & 63.98 & \underline{52.65} & \underline{62.62} & 54.08 & 58.33 & \underline{44.00} & 80.80 & \textbf{55.31} & \textbf{60.04} \\
GLM-4.1V-Thinking \cite{glm41v} \szg{9B} & 64.90 & \textbf{57.70} & \textbf{67.89} & \textbf{67.00} & \textbf{64.37} & 21.77 & 70.80 & 19.34 & 37.31 \\
MiMo-VL-RL \cite{mimovl} \szg{7B} & 59.64 & 47.14 & 56.16 & 51.59 & 53.63 & 38.28 & 77.55 & 40.39 & 52.07 \\
InternVL-3.5 \cite{internvl35} \szg{8B} & 61.73 & 49.90 & 58.15 & 54.08 & 55.96 & 42.18 & 77.95 & 49.32 & 56.48 \\
MiniCPM-V-4.5 \cite{minicpmv45} \szg{8B} & 60.41 & 47.50 & 57.16 & 53.58 & 54.66 & 41.00 & 80.85 & 44.71 & 55.52 \\
Qwen3-VL-Instruct \szg{8B} \cite{qwen3vl} & 64.18 & 49.85 & 60.04 & 49.40 & 55.87 & \textbf{45.05} & \underline{81.55} & \underline{50.16} & \underline{58.92} \\
Qwen3-VL-Thinking \cite{qwen3vl} \szg{8B} & \textbf{65.97} & 51.84 & 60.64 & 56.06 & 58.63 & 43.05 & 80.10 & 47.68 & 56.94 \\
GLM-4.6V-Flash \cite{glm41v} \szg{9B} & 64.13 & 51.48 & 61.53 & \underline{59.84} & \underline{59.25} & 42.75 & 80.45 & 48.37 & 57.19 \\
Qwen3.5 \cite{qwen3.5} \szg{9B} & 59.39 & 43.06 & 58.95 & 55.67 & 54.27 & 37.47 & 80.05 & 42.96 & 53.50 \\
LLaVA-OneVision-2 \cite{llava-onevision2} \szg{8B} & 59.90 & 48.93 & 55.67 & 50.10 & 53.65 & 41.65 & 76.35 & 46.53 & 54.84 \\
\Cbar{Middle-Size (25--38B) Open-Source MLLMs} \\
Qwen2.5-VL-Instruct \cite{2025qwen25vl} \szg{32B} & 64.08 & 49.08 & \underline{62.03} & 55.37 & 57.64 & 43.85 & 79.45 & 52.75 & 58.68 \\
Gemma-3 \cite{gemma3} \szg{27B} & \underline{64.90} & \textbf{55.10} & \textbf{66.00} & 56.16 & \textbf{60.54} & \underline{47.30} & \underline{84.70} & 55.24 & \underline{62.41} \\
InternVL3 \cite{internvl3} \szg{38B} & 64.59 & 54.29 & 60.14 & 55.47 & 58.62 & 45.60 & 81.60 & \underline{56.20} & 61.13 \\
Qwen3-VL-Instruct \cite{qwen3vl} \szg{32B} & 64.54 & 51.53 & 60.64 & 56.86 & 58.39 & 34.15 & 67.35 & 40.89 & 47.46 \\
Qwen3-VL-Thinking \cite{qwen3vl} \szg{32B} & \textbf{65.26} & 52.91 & 60.04 & \underline{58.05} & 59.06 & 45.23 & 79.30 & 50.88 & 58.47 \\
Qwen3.5 \cite{qwen3.5} \szg{27B} & 63.72 & 44.85 & 53.28 & 43.84 & 51.42 & 42.52 & 82.10 & 50.76 & 58.46 \\
Gemma-4 \cite{gemma4} \szg{31B} & 63.21 & \underline{54.74} & 60.93 & \textbf{59.94} & \underline{59.71} & \textbf{47.87} & \textbf{86.00} & \textbf{57.34} & \textbf{63.74} \\
Qwen3.6 \cite{qwen3.6-27b} \szg{27B} & 53.21 & 36.07 & 53.38 & 40.85 & 45.88 & 43.20 & 82.00 & 51.59 & 58.93 \\
\Cbar{Large-Size (72--78B) Open-Source MLLMs} \\
Qwen2.5-VL-Instruct \cite{2025qwen25vl} \szg{72B} & \underline{61.48} & \underline{48.21} & \textbf{60.14} & \textbf{58.35} & \textbf{57.05} & \underline{42.98} & \underline{81.25} & \underline{53.64} & \underline{59.29} \\
InternVL3 \cite{internvl3} \szg{78B} & \textbf{63.01} & \textbf{52.09} & \underline{57.85} & \underline{50.89} & \underline{55.96} & \textbf{45.15} & \textbf{83.15} & \textbf{57.04} & \textbf{61.78} \\
\Cbar{Proprietary MLLMs} \\
Kimi-K2.5 \cite{kimik25} & 64.39 & 53.16 & 61.43 & 60.44 & 59.85 & \underline{46.85} & \underline{84.50} & \underline{56.26} & \underline{62.54} \\
Seed-2.0-Pro \cite{seed20pro} & \textbf{67.50} & \textbf{53.67} & \textbf{65.51} & \textbf{61.23} & \textbf{61.98} & 46.43 & \textbf{84.85} & \textbf{56.39} & \textbf{62.56} \\
Qwen3.6-Plus \cite{qwen3.6-plus} & 65.31 & 53.11 & \underline{62.13} & 60.24 & 60.20 & \textbf{47.12} & 82.70 & 55.56 & 61.80 \\
GPT-5.5 \cite{gpt5} & \underline{66.17} & \underline{53.52} & 60.74 & \underline{60.83} & \underline{60.32} & 43.65 & 79.45 & 49.79 & 57.63 \\
\Cbar{Emotion-Oriented MLLMs} \\
EmoVIT \cite{cvpr2024emovit} \szg{7B} & 61.17 & 50.41 & 58.85 & \underline{54.08} & 56.13 & 46.50 & 77.75 & 50.27 & 58.17 \\
Emotion-Qwen \cite{emotion-qwen} \szg{7B} & 61.02 & 49.64 & 58.45 & 51.29 & 55.10 & 45.98 & 74.35 & 43.70 & 54.68 \\
EmoCaliber \cite{emocaliber} \szg{7B} & \underline{68.52} & \underline{52.40} & \underline{66.00} & 42.74 & \underline{57.42} & \textbf{54.60} & \textbf{82.55} & \underline{51.73} & \underline{62.96} \\
\cellcolor{oursblue}EmObserver (\textbf{Ours}) \szg{8B} & \cellcolor{oursblue}\textbf{82.86} & \cellcolor{oursblue}\textbf{61.02} & \cellcolor{oursblue}\textbf{83.60} & \cellcolor{oursblue}\textbf{59.34} & \cellcolor{oursblue}\textbf{71.70} & \cellcolor{oursblue}\underline{54.00} & \cellcolor{oursblue}\underline{81.20} & \cellcolor{oursblue}\textbf{55.05} & \cellcolor{oursblue}\textbf{63.42} \\
\bottomrule
\end{tabular}}
\vskip -0.10in
\end{table*}

\begin{figure}[t]
    \centering
    \includegraphics[width=1\linewidth]{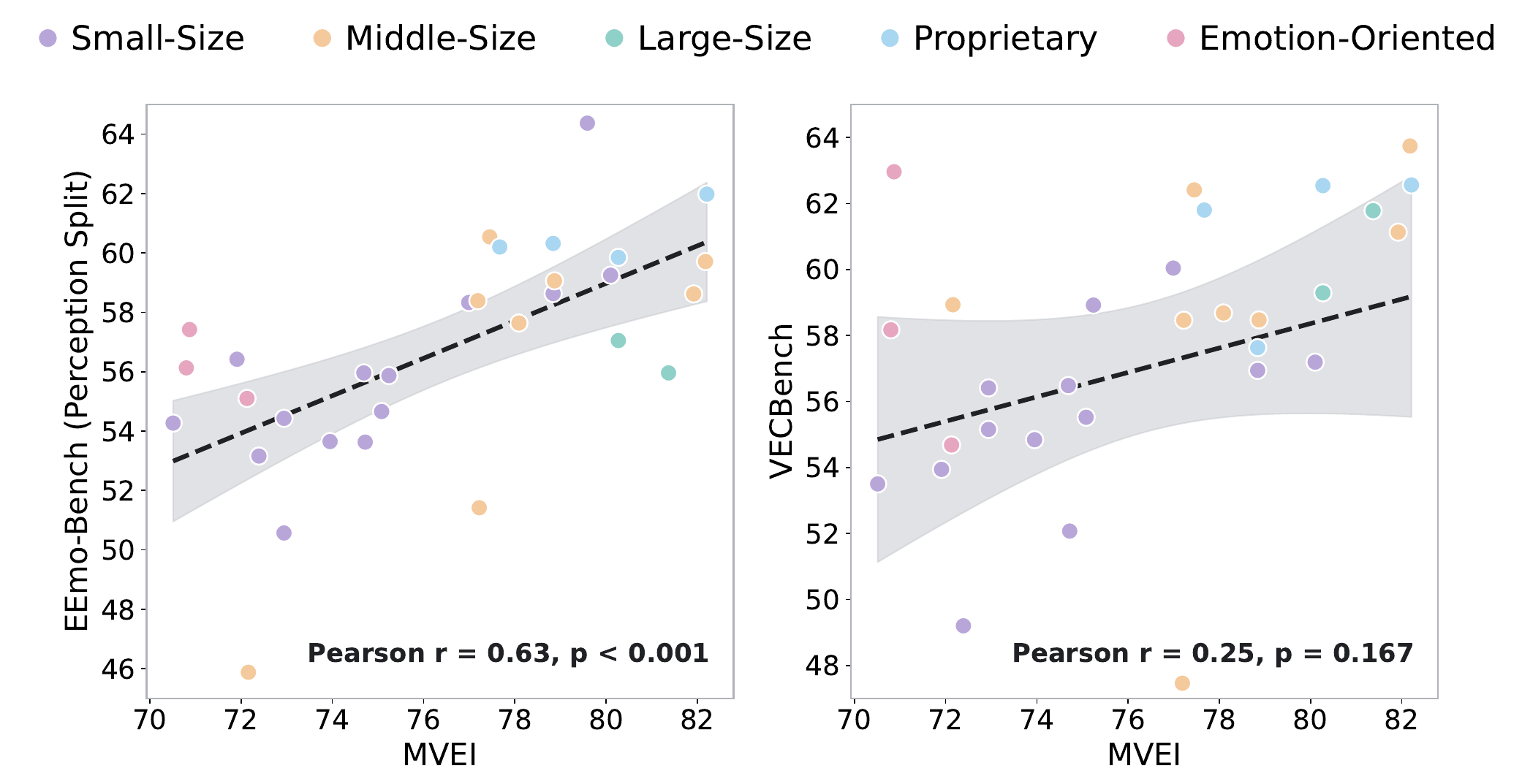}
    \vskip -0.10in
    \caption{Correlation between MVEI and existing emotion benchmarks. Each point is an MLLM (EmObserver is excluded), colored by category. The dashed line is the linear fit, and the shaded band is its 95\% confidence interval.}
    \label{fig:11}
    \vskip -0.10in
\end{figure}

\subsection{Evaluation on EEmo-Bench and VECBench}

\textit{\textbf{Model Comparison (\cref{tab:7}).}}
To compare conclusions across datasets, we further adopt two recent AICA benchmarks. EEmo-Bench \cite{mm25eemobench} combines conventional evaluation with question-answer-style formulations and additionally includes image-pair inputs; we use its perception split for evaluation. VECBench \cite{emocaliber}, in contrast, aggregates multiple traditional AICA tasks and is more closely aligned with fixed-category evaluation settings. As shown in the table, EmObserver maintains a clear advantage on EEmo-Bench, achieving an average accuracy of 71.70\%, substantially outperforming the second-best model, GLM-4.1V-Thinking, by 7.33\%. On VECBench, EmObserver ranks second among all evaluated models with 63.42\% average accuracy, surpassing all proprietary and emotion-oriented MLLMs and trailing only Gemma-4 by a small margin of 0.32\%. These results show that EmObserver's advantage is not confined to the ESJ formulation, but transfers consistently to different input formats and output requirements.

\textit{\textbf{Benchmark Comparison (\cref{fig:11}).}}
Beyond model-level comparisons, we also compare evaluation outcomes at the benchmark level to examine whether different evaluation designs lead to different rankings. As shown in \cref{fig:11}, MVEI exhibits a strong positive correlation with EEmo-Bench, with a Pearson correlation of $r=0.63$ ($p<0.001$). This high agreement cross-validates the reliability of both benchmarks, despite their distinct construction procedures and task designs. Meanwhile, the correlation is not perfect, suggesting that MVEI and EEmo-Bench still capture partially different aspects of visual emotional intelligence and thus provide complementary diagnostic value. In contrast, MVEI shows only a weaker correlation with VECBench ($r=0.25$, $p=0.167$), offering quantitative evidence of a gap between conventional AICA evaluation and modern MLLM-oriented tasks. This discrepancy further highlights the necessity of developing customized evaluation formulations such as ESJ.

\begin{table*}[]
\caption{Comparison of training recipes. Training stages: \textbf{\#1}\,=\,Cold-Start
Initialization, \textbf{\#2}\,=\,Sampling Sharpening, \textbf{\#3}\,=\,Error-Guided Refinement, \textbf{\#4}\,=\,Analysis Reinforcement. Error types: P\,=\,Perception,
D\,=\,Deviation, L\,=\,Logic, G\,=\,Grounding, O\,=\,Other.}
\label{tab:8}
\vskip -0.1in
\centering
\resizebox{1\linewidth}{!}{
\renewcommand{\arraystretch}{1.1}
\begin{tabular}{ll cccc ccccc c ccccc}
\toprule
& \multirow{2}{*}{\bf Models}
& \multicolumn{4}{c}{Training Stage}
& \multicolumn{6}{c}{MVEI: Accuracy (\%)}
& \multicolumn{5}{c}{Error Type: Proportion (\%)} \\
\cmidrule(lr){3-6} \cmidrule(lr){7-12} \cmidrule(lr){13-17}
& & \#1 & \#2 & \#3 & \#4
& SP & EI & SC & PS & Avg & Pass@8
& P & D & L & G & O \\
\midrule
\bf A  & Base Model
&  &  &  &
& 83.21 & 76.41 & 79.16 & 64.13 & 74.89 & 86.75
& 13.87 & 10.40 & 0.16 & 0.23 & 0.45 \\
\bf B  & Cold-Start Model
& \cmark &  &  &
& 74.94 & 70.93 & 76.15 & 78.32 & 74.59 & 86.52
& 8.59 & 14.06 & 1.10 & 1.65 & 0.00 \\
\bf C  & Vanilla RL Model
& \cmark & \cmark &  &
& \underline{86.47} & \textbf{83.55} & 83.88 & 83.87 & 84.09 & 86.49
& 6.12 & 7.13 & 0.16 & 2.50 & 0.00 \\
\rowcolor[HTML]{ebebeb} \bf C' & Direct RL Model
&  & \cmark &  &
& 85.21 & \underline{82.77} & 83.62 & 78.97 & 82.34 & 88.85
& 9.79 & 7.71 & 0.13 & 0.03 & 0.00 \\
\bf D  & OPSD Model
& \cmark & \cmark & \cmark &
& \textbf{87.72} & 81.72 & \underline{89.52} & \textbf{86.19} & \underline{85.55} & \textbf{91.77}
& 4.89 & 5.12 & 0.06 & 4.37 & 0.00 \\
\rowcolor[HTML]{ebebeb} \bf D' & Joint RL Model
& \cmark & \cmark &  & \cmark
& 85.46 & \textbf{83.55} & 81.78 & 82.19 & 83.02 & 89.89
& 7.91 & 8.39 & 0.62 & 0.06 & 0.00 \\
\bf E  & EmObserver
& \cmark & \cmark & \cmark & \cmark
& \underline{86.47} & \textbf{83.55} & \textbf{90.30} & \underline{86.06} & \textbf{86.23} & \underline{90.28}
& 6.64 & 6.51 & 0.36 & 0.26 & 0.00 \\
\bottomrule
\end{tabular}}
\vskip -0.10in
\end{table*}

\begin{figure}[t]
    \centering
    \includegraphics[width=1\linewidth]{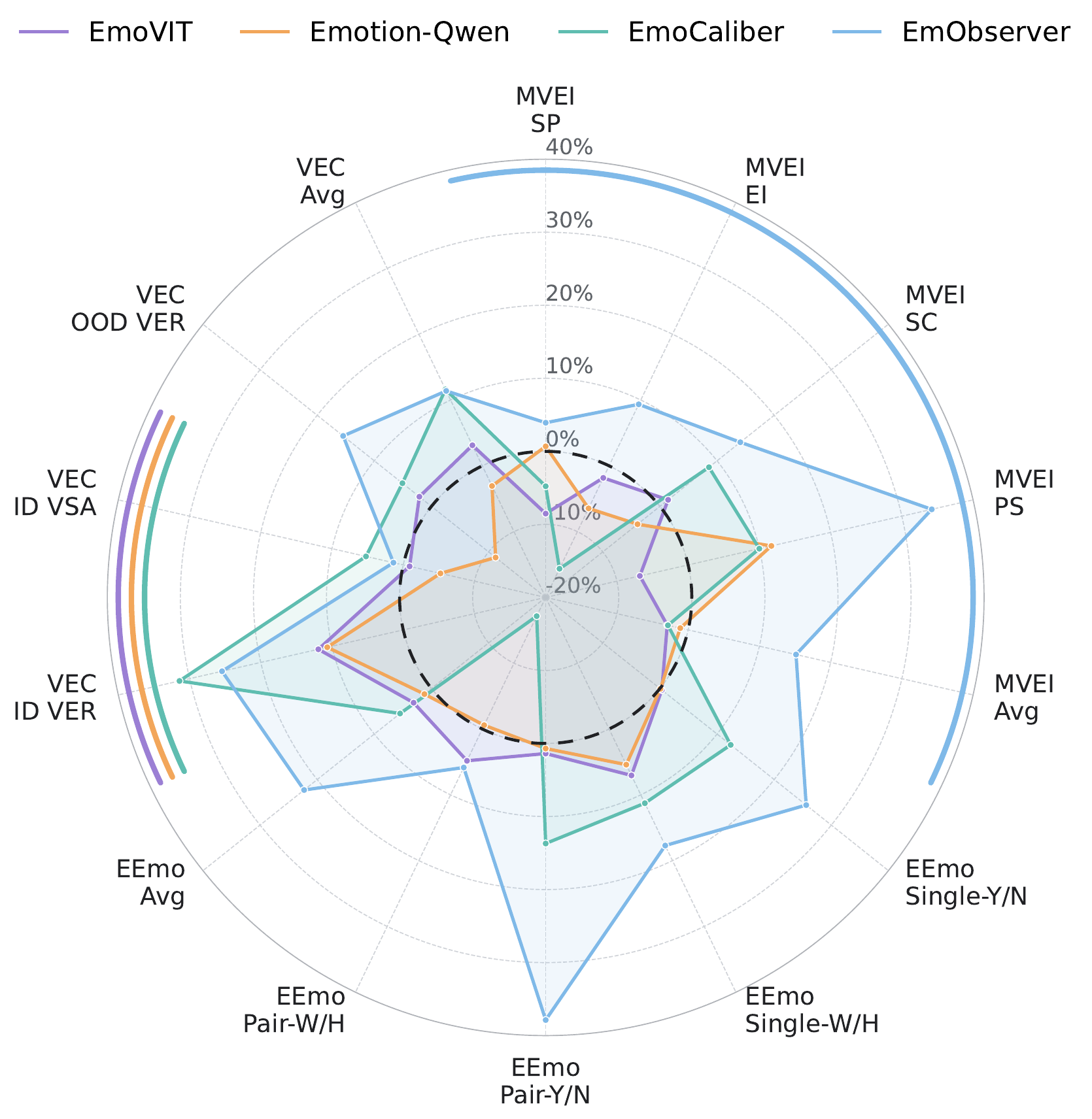}
    \vskip -0.10in
    \caption{Relative improvement of emotion-oriented MLLMs over baselines across 14 subtasks. Outer arcs denote the benchmark most aligned with each model's training objective.}
    \label{fig:12}
    \vskip -0.10in
\end{figure}

\subsection{Further Evaluation of EmObserver}

\textit{\textbf{Stage-wise Evolution (\cref{tab:8}, A$\rightarrow$B$\rightarrow$C$\rightarrow$D$\rightarrow$E).}}
To further verify the overall advantage of EmObserver, we decompose its training pipeline stage by stage. Here, A$\rightarrow$B corresponds to Cold-Start Initialization, B$\rightarrow$C to Sampling Sharpening, C$\rightarrow$D to Error-Guided Refinement, and D$\rightarrow$E to Analysis Reinforcement. The results show that almost every stage fulfills its intended role. 

The first stage equips the model with structured affective reasoning, which does not immediately translate into higher accuracy but provides a useful basis for subsequent targeted optimization. The second stage, driven by sampling-based reinforcement learning, brings the largest performance gain across the pipeline, substantially improving all four ESJ dimensions. However, its Pass@8 remains nearly unchanged (86.52\% vs. 86.49\%), indicating that sampling sharpening mainly exploits the existing response space rather than expanding it. The third stage addresses this limitation through OPSD, further improving the average accuracy from 84.09\% to 85.55\% and increasing Pass@8 to 91.77\%. This verifies the benefit of introducing privileged information to unlock higher-quality reasoning trajectories that are difficult to reach through vanilla sampling alone. From the perspective of error types, Perception and Deviation errors decrease continuously through the second and third stages, while Grounding errors gradually increase. To address this trade-off, the final Analysis Reinforcement stage explicitly encourages consistency between analysis and answer, nearly eliminating Grounding errors and yielding a better balance among error types. Overall, the multi-stage design enables EmObserver to combine high predictive accuracy with more reliable affective reasoning.

\textit{\textbf{Benefits of Cold-Start Initialization (\cref{tab:8}, A$\rightarrow$B$\rightarrow$C vs. A$\rightarrow$C').}}
Beyond reshaping the reasoning structure, Cold-Start Initialization also implicitly improves the effectiveness of the subsequent RL stage. We attribute this benefit mainly to its reduction of Perception errors, which decrease from 13.87\% in the base model to 8.59\% after cold-start training. Since perception serves as the foundation for later affective reasoning, early visual misunderstandings can easily propagate through the entire reasoning chain. By learning from high-quality structured CoT data, the model develops a more reliable perceptual basis, allowing the following RL stage to optimize from a stronger starting point. As a result, the cold-start-then-RL route (A$\rightarrow$B$\rightarrow$C) achieves 84.09\% average accuracy, outperforming direct RL from the base model (A$\rightarrow$C') at 82.34\%.

\textit{\textbf{Possibility of Joint Optimization (\cref{tab:8}, B$\rightarrow$C$\rightarrow$D$\rightarrow$E vs. B$\rightarrow$D$'$).}}
To examine the necessity of the three separate RL stages, we introduce an additional setting, denoted as B$\rightarrow$D$'$. In this setting, OPSD is removed, while the data and rewards from the second and fourth stages are merged for joint optimization. This variant achieves only 83.02\% average accuracy, significantly underperforming the adopted recipe. We attribute this degradation to the limited capability of the early-stage model to handle mixed objectives of optimizing judgement accuracy and analysis quality simultaneously. By decoupling these objectives into separate stages and using OPSD to expand the model's reasoning potential before analysis reinforcement, the proposed training recipe achieves a more effective optimization trajectory.

\begin{figure*}[t]
    \centering
    \includegraphics[width=1\linewidth]{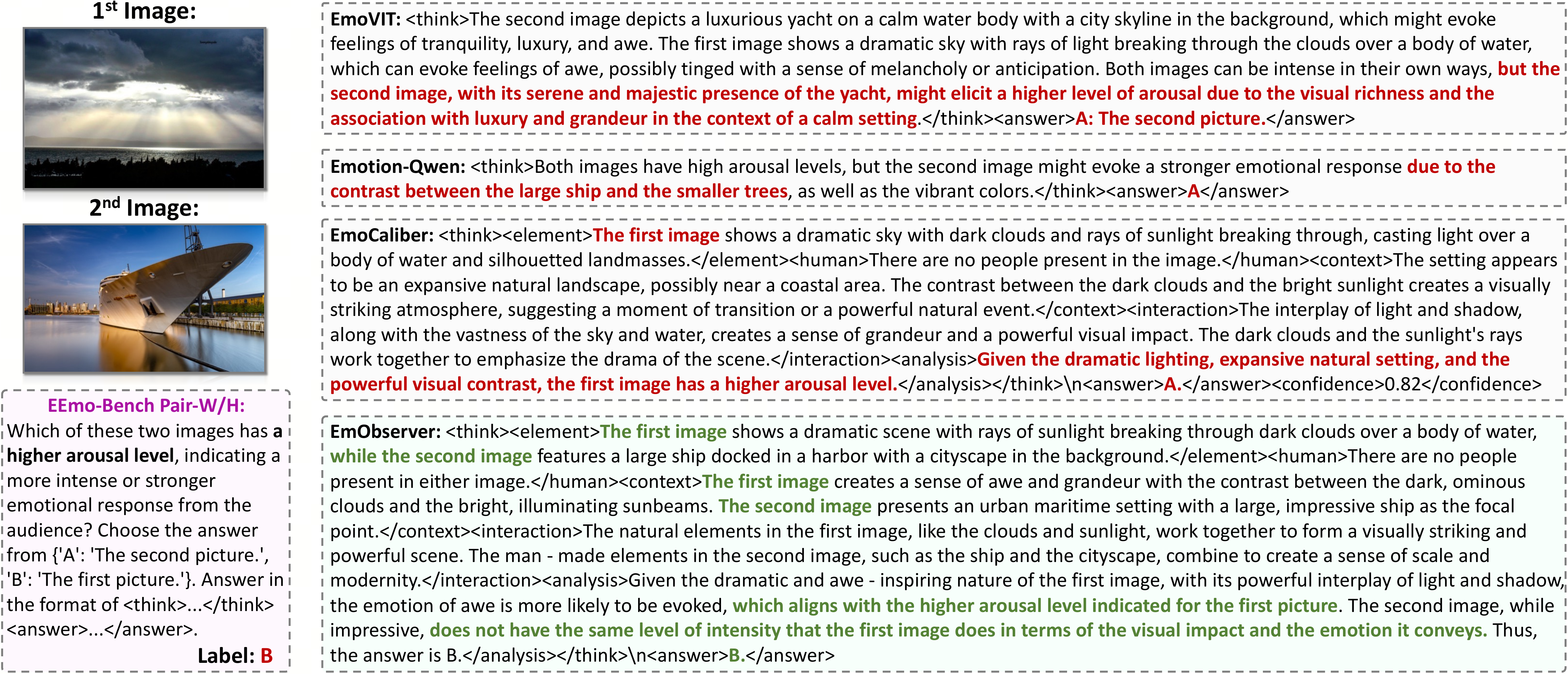}
    \vskip -0.10in
    \caption{Qualitative comparison of emotion-oriented MLLMs on a task from EEmo-Bench \cite{mm25eemobench}. The statement requires comparing two images and assessing arousal levels, which challenges generalizability beyond training targets for all MLLMs. All baselines (EmoVIT \cite{cvpr2024emovit}, Emotion-Qwen \cite{emotion-qwen}, and EmoCaliber \cite{emocaliber}) rely on superficial cues or attend to only one image and reach the wrong verdict, whereas EmObserver explicitly analyzes both images and grounds its arousal comparison in visual evidence to answer correctly.}
    \label{fig:13}
    \vskip -0.10in
\end{figure*}

\textit{\textbf{Advantages of the ESJ objective (\cref{fig:12}).}}
The preceding experiments demonstrate the overall superiority of EmObserver. Here we further visualize the relative gains of each model after emotion-oriented optimization. As shown in the figure, EmObserver consistently outperforms its base model across all task dimensions, whereas prior emotion-oriented MLLMs improve on their targeted tasks but struggle to generalize to other dimensions. This provides direct evidence for the advantage of ESJ as an optimization objective. By expanding the input space while reducing ambiguity in the output space, ESJ avoids forcing potentially plausible answers apart, yet still preserves sufficient task depth and breadth for transferable affective learning. Notably, prior models are built upon Qwen2.5-VL-Instruct-7B \cite{2025qwen25vl} as the baseline (EmoVIT is a reproduced version for fair comparison), while EmObserver, based on a stronger Qwen3-VL-Thinking-8B, achieves stronger and more consistent relative gains, further underscoring the effectiveness of ESJ-targeted optimization.

\textit{\textbf{Case Visualization (\cref{fig:13}).}}
To more intuitively demonstrate the expertise of EmObserver, we further visualize a representative case. The task requires the model to compare the arousal levels evoked by two images, a task format that never appears in the training data. Nevertheless, EmObserver exhibits strong analytical ability: it jointly compares the visual content and the elicited emotional responses of the two images, and finally arrives at the correct answer.

\section{Conclusion}
In this paper, we revisit visual emotional intelligence in MLLMs and identify the mismatch between conventional AICA paradigms and the open-ended, instruction-driven nature of modern MLLMs. To address this issue, we introduce the ESJ task, which reformulates AICA as a controlled yet expressive statement-verification problem. Built upon ESJ, we develop the INSETS pipeline and introduce the large-scale INSETS-462k corpus and the human-refined MVEI benchmark. We further propose EmObserver, an emotion-oriented MLLM optimized primarily on the ESJ task through a multi-stage recipe. Extensive evaluations show that MVEI exposes substantial capability gaps among current MLLMs, while EmObserver achieves strong performance, robust generalization, and faithful affective reasoning across multiple benchmarks. Overall, these findings suggest that visual emotional intelligence remains underexplored in general-purpose MLLMs and requires dedicated task formulations, data construction, and optimization strategies. We hope this work provides a useful foundation and inspires future research toward MLLMs with more reliable, nuanced, and human-aligned visual emotional intelligence.

\bibliographystyle{IEEEtran}
\bibliography{main}

@String(CVPR= {IEEE Conf. Comput. Vis. Pattern Recog.})

@String(ICCV= {Int. Conf. Comput. Vis.})

@String(ECCV= {Eur. Conf. Comput. Vis.})

@String(ACMMM= {ACM Int. Conf. Multimedia})

@String(ICLR = {Int. Conf. Learn. Represent.})

@String(AAAI = {AAAI})

@String(CVPR  = {CVPR})

@String(ICCV  = {ICCV})

@String(ECCV  = {ECCV})

@String(ACMMM = {ACM MM})

@String(ICLR  = {ICLR})

@article{tpami2022review,
  author       = {Sicheng Zhao and
                  Xingxu Yao and
                  Jufeng Yang and
                  Guoli Jia and
                  Guiguang Ding and
                  Tat{-}Seng Chua and
                  Bj{\"{o}}rn W. Schuller and
                  Kurt Keutzer},
  title        = {Affective Image Content Analysis: Two Decades Review and New Perspectives},
  journal      = {Trans. Pattern Anal. Mach. Intell.},
  volume       = {44},
  number       = {10},
  pages        = {6729--6751},
  year         = {2022},
}

@misc{arxiv2023mmbigbench,
  author       = {Xiaocui Yang and
                  Wenfang Wu and
                  Shi Feng and
                  Ming Wang and
                  Daling Wang and
                  others},
  title        = {MM-BigBench: Evaluating Multimodal Models on Multimodal Content Comprehension Tasks},
  year         = {2023},
  eprint       = {2310.09036},
  archivePrefix={arXiv},
}

@inproceedings{icml2021clip,
  author       = {Alec Radford and
                  Jong Wook Kim and
                  Chris Hallacy and
                  Aditya Ramesh and
                  Gabriel Goh and
                  et. al.},
  title        = {Learning Transferable Visual Models From Natural Language Supervision},
  booktitle    = {ICML},
  series       = {Proceedings of Machine Learning Research},
  volume       = {139},
  pages        = {8748--8763},
  year         = {2021},
}

@inproceedings{mm2016opinion,
  author       = {Sicheng Zhao and
                  Hongxun Yao and
                  Yue Gao and
                  Rongrong Ji and
                  Wenlong Xie and
                  Xiaolei Jiang and
                  Tat{-}Seng Chua},
  title        = {Predicting Personalized Emotion Perceptions of Social Images},
  booktitle    = {ACMMM},
  pages        = {1385--1394},
  year         = {2016},
}

@article{schutte2001decision,
  title={Emotional intelligence and interpersonal relations},
  author={Schutte, Nicola S and Malouff, John M and Bobik, Chad and Coston, Tracie D and Greeson, Cyndy and Jedlicka, Christina and Rhodes, Emily and Wendorf, Greta},
  journal={The Journal of Social Psychology},
  volume={141},
  number={4},
  pages={523--536},
  year={2001},
}

@inproceedings{cvpr2024emovit,
  author       = {Hongxia Xie and
                  Chu{-}Jun Peng and
                  Yu{-}Wen Tseng and
                  Hung{-}Jen Chen and
                  Chan{-}Feng Hsu and
                  Hong{-}Han Shuai and
                  Wen{-}Huang Cheng},
  title        = {EmoVIT: Revolutionizing Emotion Insights with Visual Instruction Tuning},
  booktitle    = {CVPR},
  pages        = {26586--26595},
  year         = {2024},
}

@inproceedings{nips2024emollama,
  author       = {Zebang Cheng and
                  Zhi{-}Qi Cheng and
                  Jun{-}Yan He and
                  Kai Wang and
                  Yuxiang Lin and
                  Zheng Lian and
                  Xiaojiang Peng and
                  Alexander G. Hauptmann},
  title        = {Emotion-LLaMA: Multimodal Emotion Recognition and Reasoning with Instruction Tuning},
  booktitle    = {NeurIPS},
  year         = {2024},
}

@inproceedings{cvpr2022mdan,
  author       = {Liwen Xu and
                  Zhengtao Wang and
                  Bin Wu and
                  Simon Lui},
  title        = {{MDAN:} Multi-level Dependent Attention Network for Visual Emotion
                  Analysis},
  booktitle    = {CVPR},
  pages        = {9469--9478},
  publisher    = {{IEEE}},
  year         = {2022},
}

@inproceedings{cvpr2017emotic,
  author       = {Ronak Kosti and
                  Jos{\'{e}} M. {\'{A}}lvarez and
                  Adri{\`{a}} Recasens and
                  {\`{A}}gata Lapedriza},
  title        = {{EMOTIC:} Emotions in Context Dataset},
  booktitle    = {CVPR Workshops},
  pages        = {2309--2317},
  year         = {2017},
}

@inproceedings{aaai2016fi,
  author       = {Quanzeng You and
                  Jiebo Luo and
                  Hailin Jin and
                  Jianchao Yang},
  title        = {Building a Large Scale Dataset for Image Emotion Recognition: The
                  Fine Print and The Benchmark},
  booktitle    = {AAAI},
  pages        = {308--314},
  year         = {2016},
}

@inproceedings{cvpr2023affection,
  author       = {Panos Achlioptas and
                  Maks Ovsjanikov and
                  Leonidas J. Guibas and
                  Sergey Tulyakov},
  title        = {Affection: Learning Affective Explanations for Real-World Visual Data},
  booktitle    = {CVPR},
  pages        = {6641--6651},
  year         = {2023},
}

@inproceedings{mm2010abstract,
  author       = {Jana Machajdik and
                  Allan Hanbury},
  title        = {Affective Image Classification Using Features Inspired by Psychology
                  and Art Theory},
  booktitle    = {ACMMM},
  pages        = {83--92},
  year         = {2010},
}

@inproceedings{iccv2023emoset,
  author       = {Jingyuan Yang and
                  Qirui Huang and
                  Tingting Ding and
                  Dani Lischinski and
                  Daniel Cohen{-}Or and
                  Hui Huang},
  title        = {EmoSet: {A} Large-scale Visual Emotion Dataset with Rich Attributes},
  booktitle    = {ICCV},
  pages        = {20326--20337},
  year         = {2023},
}

@inproceedings{cvpr2021artemis,
  author       = {Panos Achlioptas and
                  Maks Ovsjanikov and
                  Kilichbek Haydarov and
                  Mohamed Elhoseiny and
                  Leonidas J. Guibas},
  title        = {ArtEmis: Affective Language for Visual Art},
  booktitle    = {CVPR},
  pages        = {11569--11579},
  year         = {2021},
}

@inproceedings{wacv2019lucfer,
  author       = {Pooyan Balouchian and
                  Marjaneh Safaei and
                  Hassan Foroosh},
  title        = {{LUCFER:} {A} Large-Scale Context-Sensitive Image Dataset for Deep
                  Learning of Visual Emotions},
  booktitle    = {WACV},
  pages        = {1645--1654},
  year         = {2019},
}

@article{barrett2011context,
  title={Context in Emotion Perception},
  author={Barrett, Lisa Feldman and Mesquita, Batja and Gendron, Maria},
  journal={Current Directions in Psychological Science},
  volume={20},
  number={5},
  pages={286--290},
  year={2011},
}

@inproceedings{cvpr2017crowdscource,
  author       = {Shan Li and
                  Weihong Deng and
                  Junping Du},
  title        = {Reliable Crowdsourcing and Deep Locality-Preserving Learning for Expression
                  Recognition in the Wild},
  booktitle    = {CVPR},
  pages        = {2584--2593},
  year         = {2017},
}

@article{arxiv2023gpt4,
  author       = {OpenAI},
  title        = {{GPT-4} Technical Report},
  journal      = {CoRR},
  volume       = {abs/2303.08774},
  year         = {2023},
}

@article{hurst2024gpt4o,
  title={Gpt-4o system card},
  author={Hurst, Aaron and Lerer, Adam and Goucher, Adam P and Perelman, Adam and Ramesh, Aditya and Clark, Aidan and Ostrow, AJ and Welihinda, Akila and Hayes, Alan and Radford, Alec and others},
  journal={arXiv preprint arXiv:2410.21276},
  year={2024}
}

@article{mikels2005emotional,
  title={Emotional Category Data on Images from the International Affective Picture System},
  author={Mikels, Joseph A and Fredrickson, Barbara L and Larkin, Gregory R and Lindberg, Casey M and Maglio, Sam J and Reuter-Lorenz, Patricia A},
  journal={Behavior Research Methods},
  volume={37},
  pages={626--630},
  year={2005},
}

@book{parrott2001emotions,
  title={Emotions in Social Psychology: Essential Readings},
  author={Parrott, W Gerrod},
  year={2001},
  publisher={Psychology Press}
}

@inproceedings{eccv2022s2ver,
  author       = {Guoli Jia and
                  Jufeng Yang},
  title        = {S\({}^{\mbox{2}}\)-VER: Semi-supervised Visual Emotion Recognition},
  booktitle    = {ECCV},
  volume       = {13697},
  pages        = {493--509},
  year         = {2022},
}

@inproceedings{cvpr2023prob,
  author       = {Tinglei Feng and
                  Jiaxuan Liu and
                  Jufeng Yang},
  title        = {Probing Sentiment-Oriented PreTraining Inspired by Human Sentiment
                  Perception Mechanism},
  booktitle    = {CVPR},
  pages        = {2850--2860},
  year         = {2023},
}

@inproceedings{mm24pacl,
  author       = {Daiqing Wu and
                  Dongbao Yang and
                  Yu Zhou and
                  Can Ma},
  title        = {Bridging Visual Affective Gap: Borrowing Textual Knowledge by Learning
                  from Noisy Image-Text Pairs},
  booktitle    = {ACMMM},
  pages        = {602--611},
  year         = {2024},
}

@inproceedings{agi2020condition,
  author       = {Yoshihiro Maruyama},
  title        = {The Conditions of Artificial General Intelligence: Logic, Autonomy,
                  Resilience, Integrity, Morality, Emotion, Embodiment, and Embeddedness},
  booktitle    = {AGI},
  series       = {Lecture Notes in Computer Science},
  volume       = {12177},
  pages        = {242--251},
  year         = {2020},
}

@inproceedings{acl2024emobench,
  author       = {Sahand Sabour and
                  Siyang Liu and
                  Zheyuan Zhang and
                  June M. Liu and
                  Jinfeng Zhou and
                  others},
  title        = {EmoBench: Evaluating the Emotional Intelligence of Large Language
                  Models},
  booktitle    = {ACL},
  pages        = {5986--6004},
  year         = {2024},
}

@inproceedings{eccv2024faba,
  author       = {Yifan Li and
                  Anh Dao and
                  Wentao Bao and
                  Zhen Tan and
                  Tianlong Chen and
                  Huan Liu and
                  Yu Kong},
  title        = {Facial Affective Behavior Analysis with Instruction Tuning},
  booktitle    = {ECCV},
  series       = {Lecture Notes in Computer Science},
  volume       = {15076},
  pages        = {165--186},
  year         = {2024},
}

@article{arxiv2024hallucination,
  author       = {Zechen Bai and
                  Pichao Wang and
                  Tianjun Xiao and
                  Tong He and
                  Zongbo Han and
                  Zheng Zhang and
                  Mike Zheng Shou},
  title        = {Hallucination of Multimodal Large Language Models: {A} Survey},
  journal      = {CoRR},
  volume       = {abs/2404.18930},
  year         = {2024},
}

@inproceedings{eccv2024safebench,
  author       = {Xin Liu and
                  Yichen Zhu and
                  Jindong Gu and
                  Yunshi Lan and
                  Chao Yang and
                  Yu Qiao},
  title        = {MM-SafetyBench: {A} Benchmark for Safety Evaluation of Multimodal
                  Large Language Models},
  booktitle    = {ECCV},
  series       = {Lecture Notes in Computer Science},
  volume       = {15114},
  pages        = {386--403},
  year         = {2024},
}

@article{spm2006gap,
  title={Extracting Moods from Pictures and Sounds: Towards Truly Personalized TV},
  author={Hanjalic, Alan},
  journal={IEEE Signal Processing Magazine},
  volume={23},
  number={2},
  pages={90--100},
  year={2006},
}

@inproceedings{cvpr2024llava_next,
  author       = {Haotian Liu and
                  Chunyuan Li and
                  Yuheng Li and
                  Yong Jae Lee},
  title        = {Improved Baselines with Visual Instruction Tuning},
  booktitle    = {CVPR},
  pages        = {26286--26296},
  year         = {2024},
}

@article{arxiv2024mantis,
  author       = {Dongfu Jiang and
                  Xuan He and
                  Huaye Zeng and
                  Cong Wei and
                  Max Ku and
                  Qian Liu and
                  Wenhu Chen},
  title        = {{MANTIS:} Interleaved Multi-Image Instruction Tuning},
  journal      = {CoRR},
  volume       = {abs/2405.01483},
  year         = {2024},
}

@article{arxiv2024mplugowl3,
  author       = {Jiabo Ye and
                  Haiyang Xu and
                  Haowei Liu and
                  Anwen Hu and
                  Ming Yan and
                  others},
  title        = {mPLUG-Owl3: Towards Long Image-Sequence Understanding in Multi-Modal
                  Large Language Models},
  journal      = {CoRR},
  volume       = {abs/2408.04840},
  year         = {2024},
}

@article{arxiv2024idefics3,
  author       = {Hugo Lauren{\c{c}}on and
                  Andr{\'{e}}s Marafioti and
                  Victor Sanh and
                  L{\'{e}}o Tronchon},
  title        = {Building and Better Understanding Vision-Language Models: Insights
                  and Future Directions},
  journal      = {CoRR},
  volume       = {abs/2408.12637},
  year         = {2024},
}

@article{arxiv2024qwen2vl,
  author       = {Peng Wang and
                  Shuai Bai and
                  Sinan Tan and
                  Shijie Wang and
                  Zhihao Fan and
                  others},
  title        = {Qwen2-VL: Enhancing Vision-Language Model's Perception of the World at Any Resolution},
  journal      = {CoRR},
  volume       = {abs/2409.12191},
  year         = {2024},
}

@article{arxiv2024llama3,
  author       = {Abhimanyu Dubey and
                  Abhinav Jauhri and
                  Abhinav Pandey and
                  Abhishek Kadian and
                  Ahmad Al{-}Dahle and
                  others},
  title        = {The Llama 3 Herd of Models},
  journal      = {CoRR},
  volume       = {abs/2407.21783},
  year         = {2024},
}

@article{arxiv2024molmo,
  author       = {Matt Deitke and
                  Christopher Clark and
                  Sangho Lee and
                  Rohun Tripathi and
                  Yue Yang and
                  others},
  title        = {Molmo and PixMo: Open Weights and Open Data for State-of-the-Art Multimodal
                  Models},
  journal      = {CoRR},
  volume       = {abs/2409.17146},
  year         = {2024},
}

@article{arxiv2024phi3,
  author       = {Marah I Abdin and
                  Sam Ade Jacobs and
                  Ammar Ahmad Awan and
                  Jyoti Aneja and
                  Ahmed Awadallah and
                  others},
  title        = {Phi-3 Technical Report: {A} Highly Capable Language Model Locally
                  on Your Phone},
  journal      = {CoRR},
  volume       = {abs/2404.14219},
  year         = {2024},
}

@article{arxiv2024intervl25,
  author       = {Zhe Chen and
                  Weiyun Wang and
                  Yue Cao and
                  Yangzhou Liu and
                  Zhangwei Gao and
                  others},
  title        = {Expanding Performance Boundaries of Open-Source Multimodal Models
                  with Model, Data, and Test-Time Scaling},
  journal      = {CoRR},
  volume       = {abs/2412.05271},
  year         = {2024},
}

@misc{2023opencompass,
    title={OpenCompass: A Universal Evaluation Platform for Foundation Models},
    author={OpenCompass Contributors},
    howpublished = {\url{https://github.com/open-compass/opencompass}},
    year={2023}
}

@inproceedings{nips2023instructblip,
  author       = {Wenliang Dai and
                  Junnan Li and
                  Dongxu Li and
                  Anthony Meng Huat Tiong and
                  Junqi Zhao and
                  others},
  title        = {InstructBLIP: Towards General-purpose Vision-Language Models with
                  Instruction Tuning},
  booktitle    = {NeurIPS},
  year         = {2023},
}

@article{arxiv2024deepseekvl,
  author       = {Haoyu Lu and
                  Wen Liu and
                  Bo Zhang and
                  Bingxuan Wang and
                  Kai Dong and
                  others},
  title        = {DeepSeek-VL: Towards Real-World Vision-Language Understanding},
  journal      = {CoRR},
  volume       = {abs/2403.05525},
  year         = {2024},
}

@article{arxiv2024minicpm,
  author       = {Yuan Yao and
                  Tianyu Yu and
                  Ao Zhang and
                  Chongyi Wang and
                  Junbo Cui and
                  others},
  title        = {MiniCPM-V: {A} {GPT-4V} Level {MLLM} on Your Phone},
  journal      = {CoRR},
  volume       = {abs/2408.01800},
  year         = {2024},
}

@misc{2025qwen25vl,
    title = {Qwen2.5-VL},
    howpublished = {\url{https://qwenlm.github.io/blog/qwen2.5-vl/}},
    author = {Qwen Team},
    year = {2025}
}

@inproceedings{cvpr2020web,
  title={Learning Visual Emotion Representations from Web Data},
  author={Wei, Zijun and Zhang, Jianming and Lin, Zhe and Lee, Joon-Young and Balasubramanian, Niranjan and Hoai, Minh and Samaras, Dimitris},
  booktitle={CVPR},
  pages={13106--13115},
  year={2020}
}

@article{pieee2023label-effic,
  title={Toward Label-efficient Emotion and Sentiment Analysis},
  author={Zhao, Sicheng and Hong, Xiaopeng and Yang, Jufeng and Zhao, Yanyan and Ding, Guiguang},
  journal={Proceedings of the IEEE},
  volume={111},
  number={10},
  pages={1159--1197},
  year={2023},
}

@InProceedings{2024eibench,
    author    = {Lin, Yuxiang and Sun, Jingdong and Cheng, Zhi-Qi and Wang, Jue and Liang, Haomin and Cheng, Zebang and Dong, Yifei and He, Jun-Yan and Peng, Xiaojiang and Hua, Xian-Sheng},
    title     = {Why We Feel: Breaking Boundaries in Emotional Reasoning with Multimodal Large Language Models},
    booktitle = {CVPR Workshops},
    month     = {June},
    year      = {2025},
    pages     = {5205-5215}
}

@inproceedings{icml2025icl,
  title={An Empirical Study on Configuring In-Context Learning Demonstrations for Unleashing MLLMs' Sentimental Perception Capability},
  author={Wu, Daiqing and Yang, Dongbao and Zhao, Sicheng and Ma, Can and Zhou, Yu},
  booktitle={ICML},
  year={2025},
}

@article{individual2004,
  title={Individual differences in emotion processing},
  author={Hamann, Stephan and Canli, Turhan},
  journal={Current opinion in neurobiology},
  volume={14},
  number={2},
  pages={233--238},
  year={2004},
  publisher={Elsevier}
}

@article{wieser2012faces,
  title={Faces in context: A review and systematization of contextual influences on affective face processing},
  author={Wieser, Matthias J and Brosch, Tobias},
  journal={Frontiers in psychology},
  volume={3},
  pages={471},
  year={2012},
  publisher={Frontiers Media SA}
}

@article{hu2025emobench,
  title={Emobench-m: Benchmarking emotional intelligence for multimodal large language models},
  author={Hu, He and Zhou, Yucheng and You, Lianzhong and Xu, Hongbo and Wang, Qianning and Lian, Zheng and Yu, Fei Richard and Ma, Fei and Cui, Laizhong},
  journal={arXiv preprint arXiv:2502.04424},
  year={2025}
}

@article{1980circumplex,
  title={A Circumplex Model of Affect},
  author={Russell, James A},
  journal={Journal of Personality and Social Psychology},
  volume={39},
  number={6},
  pages={1161},
  year={1980},
  publisher={American Psychological Association}
}

@article{1971constants,
  title={Constants across Cultures in the Face and Emotion},
  author={Ekman, Paul and Friesen, Wallace V},
  journal={Journal of Personality and Social Psychology},
  volume={17},
  number={2},
  pages={124},
  year={1971},
  publisher={American Psychological Association}
}

@article{2017cog+emo,
  title={Emotion Perception from a Componential Perspective},
  author={Shuman, Vera and Clark-Polner, Elizabeth and Meuleman, Ben and Sander, David and Scherer, Klaus R},
  journal={Cognition and Emotion},
  volume={31},
  number={1},
  pages={47--56},
  year={2017},
  publisher={Taylor \& Francis}
}

@article{shao2024deepseekmath,
  title={Deepseekmath: Pushing the limits of mathematical reasoning in open language models},
  author={Shao, Zhihong and Wang, Peiyi and Zhu, Qihao and Xu, Runxin and Song, Junxiao and Bi, Xiao and Zhang, Haowei and Zhang, Mingchuan and Li, YK and Wu, Yang and others},
  journal={arXiv preprint arXiv:2402.03300},
  year={2024}
}

@misc{qwen3.5,
    title  = {{Qwen3.5}: Towards Native Multimodal Agents},
    author = {{Qwen Team}},
    year   = {2026},
    url    = {https://qwen.ai/blog?id=qwen3.5}
}

@misc{qwen3.6-27b,
    title  = {{Qwen3.6-27B}: Flagship-Level Coding in a {27B} Dense Model},
    author = {{Qwen Team}},
    year   = {2026},
    url    = {https://qwen.ai/blog?id=qwen3.6-27b}
}

@inproceedings{minicpmv45,
  title={Minicpm-v 4.5: Cooking Efficient MLLMs via Architecture, Data, and Training Recipe},
  author={Yu, Tianyu and Wang, Zefan and Wang, Chongyi and Huang, Fuwei and Ma, Wenshuo and He, Zhihui and Cai, Tianchi and Chen, Weize and Huang, Yuxiang and Zhao, Ranchi and others},
  booktitle = {CVPR},
  pages={11704--11715},
  year={2026}
}

@article{internvl3,
  title={InternVL3: Exploring Advanced Training and Test-Time Recipes for Open-Source Multimodal Models},
  author={Zhu, Jinguo and Wang, Weiyun and Chen, Zhe and Liu, Zhaoyang and Ye, Shenglong and Gu, Lixin and Tian, Hao and Duan, Yuchen and Su, Weijie and Shao, Jie and others},
  journal={arXiv preprint arXiv:2504.10479},
  year={2025}
}

@article{glm41v,
  title={GLM-4.1 V-Thinking: Towards Versatile Multimodal Reasoning with Scalable Reinforcement Learning},
  author={Hong, Wenyi and Yu, Wenmeng and Gu, Xiaotao and Wang, Guo and Gan, Guobing and Tang, Haomiao and Cheng, Jiale and Qi, Ji and Ji, Junhui and Pan, Lihang and others},
  journal={arXiv preprint arXiv:2507.01006,},
  year={2025}
}

@article{mimovl,
  title={MiMo-VL Technical Report},
  author={Xiaomi LLM-Core Team},
  journal={arXiv preprint arXiv:2506.03569,},
  year={2025}
}

@article{internvl35,
  title={Internvl3. 5: Advancing Open-Source Multimodal Models in Versatility, Reasoning, and Efficiency},
  author={Wang, Weiyun and Gao, Zhangwei and Gu, Lixin and Pu, Hengjun and Cui, Long and Wei, Xingguang and Liu, Zhaoyang and Jing, Linglin and Ye, Shenglong and Shao, Jie and others},
  journal={arXiv preprint arXiv:2508.18265},
  year={2025}
}

@article{qwen3vl,
  title={Qwen3-VL Technical Report},
  author={Bai, Shuai and Cai, Yuxuan and Chen, Ruizhe and Chen, Keqin and Chen, Xionghui and Cheng, Zesen and Deng, Lianghao and Ding, Wei and Gao, Chang and Ge, Chunjiang and others},
  journal={arXiv preprint arXiv:2511.21631},
  year={2025}
}

@article{gemma3,
  title={Gemma 3 Technical Report},
  author={Gemma Team},
  journal={arXiv preprint arXiv:2503.19786},
  year={2025}
}

@misc{gemma4,
    title  = {Gemma 4: Byte for byte, the most capable open models},
    author = {Gemma Team},
    year   = {2026},
    url    ={https://blog.google/innovation-and-ai/technology/developers-tools/gemma-4/}
}

@article{llava-onevision2,
  title={LLaVA-OneVision-2: Towards Next-Generation Perceptual Intelligence},
  author={An, Xiang and Xie, Yin and Tang, Feilong and Yan, Yunyao and Tan, Huajie and Zhu, Didi and Chen, Changrui and Zhao, Xiuwei and Qin, Bin and Yang, Kaicheng and others},
  journal={arXiv preprint arXiv:2605.25979},
  year={2026}
}

@article{kimik25,
  title={Kimi K2.5: Visual Agentic Intelligence},
  author={Kimi Team},
  journal={arXiv preprint arXiv:2602.02276},
  year={2026}
}

@misc{seed20pro,
    title  = {Seed 2.0 Official Launch},
    author = {{Seed Team}},
    year   = {2026},
    url    = {https://seed.bytedance.com/en/blog/seed-2-0-official-launch}
}

@misc{qwen3.6-plus,
    title  = {Qwen3.6-Plus: Towards Real World Agents},
    author = {{Qwen Team}},
    year   = {2026},
    url    = {https://qwen.ai/blog?id=qwen3.6}
}

@article{gpt5,
  title={OpenAI GPT-5 System Card},
  author={Singh, Aaditya and Fry, Adam and Perelman, Adam and Tart, Adam and Ganesh, Adi and El-Kishky, Ahmed and McLaughlin, Aidan and Low, Aiden and Ostrow, AJ and Ananthram, Akhila and others},
  journal={arXiv preprint arXiv:2601.03267},
  year={2026}
}

@article{emotion-qwen,
  title={Emotion-Qwen: A Unified Framework for Emotion and Vision Understanding},
  author={Huang, Dawei and Li, Qing and Yan, Chuan and Cheng, Zebang and Han, Zihao and Huang, Yurong and Li, Xiang and Li, Bin and Wang, Xiaohui and Lian, Zheng and others},
  journal={arXiv preprint arXiv:2505.06685},
  year={2025}
}

@article{emocaliber,
  title={EmoCaliber: Advancing Reliable Visual Emotion Comprehension via Confidence Verbalization and Calibration},
  author={Wu, Daiqing and Yang, Dongbao and Ma, Can and Zhou, Yu},
  journal={Pattern Recognition},
  pages={113716},
  year={2026},
  publisher={Elsevier}
}

@inproceedings{mm25eemobench,
  author       = {Lancheng Gao and
                  Ziheng Jia and
                  Yunhao Zeng and
                  Wei Sun and
                  Yiming Zhang and
                  Wei Zhou and
                  Guangtao Zhai and
                  Xiongkuo Min},
  title        = {EEmo-Bench: {A} Benchmark for Multi-modal Large Language Models on
                  Image Evoked Emotion Assessment},
  booktitle    = {ACMMM},
  pages        = {7064--7073},
  year         = {2025},
}

@article{mer-llm,
  title={Multimodal Emotion Recognition with Large Language Models},
  author={Zhang, Hongrui and Wu, Daiqing and Li, Yangyang and Liu, Kuien and Wang, Yuhui and Zhou, Yu and Zhao, Sicheng},
  journal={arXiv preprint arXiv:2605.21239},
  year={2026}
}

@inproceedings{wscnet,
  author       = {Jufeng Yang and
                  Dongyu She and
                  Yu{-}Kun Lai and
                  Paul L. Rosin and
                  Ming{-}Hsuan Yang},
  title        = {Weakly Supervised Coupled Networks for Visual Sentiment Analysis},
  booktitle    = {CVPR},
  pages        = {7584--7592},
  year         = {2018},
}

@inproceedings{padnet,
  author       = {Sicheng Zhao and
                  Zizhou Jia and
                  Hui Chen and
                  Leida Li and
                  Guiguang Ding and
                  Kurt Keutzer},
  title        = {PDANet: Polarity-Consistent Deep Attention Network for Fine-grained
                  Visual Emotion Regression},
  booktitle    = {ACMMM},
  pages        = {192--201},
  year         = {2019},
}

@inproceedings{artemisv2,
  author       = {Youssef Mohamed and
                  Faizan Farooq Khan and
                  Kilichbek Haydarov and
                  Mohamed Elhoseiny},
  title        = {It is Okay to Not Be Okay: Overcoming Emotional Bias in Affective
                  Image Captioning by Contrastive Data Collection},
  booktitle    = {CVPR},
  pages        = {21231--21240},
  year         = {2022},
}

@inproceedings{iccv2019zero-shot1,
  author       = {Chi Zhan and
                  Dongyu She and
                  Sicheng Zhao and
                  Ming{-}Ming Cheng and
                  Jufeng Yang},
  title        = {Zero-Shot Emotion Recognition via Affective Structural Embedding},
  booktitle    = {ICCV},
  pages        = {1151--1160},
  year         = {2019},
}

@inproceedings{mmasia2022zero-shot2,
  author       = {Yingrui Ye and
                  Yuya Moroto and
                  Keisuke Maeda and
                  Takahiro Ogawa and
                  Miki Haseyama},
  title        = {Affective Embedding Framework with Semantic Representations from Tweets
                  for Zero-Shot Visual Sentiment Prediction},
  booktitle    = {ACMMMAsia},
  pages        = {6:1--6:7},
  year         = {2022},
}

@article{eemologic,
  title={EEmo-Logic: A Unified Dataset and Multi-Stage Framework for Comprehensive Image-Evoked Emotion Assessment},
  author={Gao, Lancheng and Jia, Ziheng and Xing, Zixuan and Sun, Wei and Duan, Huiyu and Zhai, Guangtao and Min, Xiongkuo},
  journal={arXiv preprint arXiv:2602.01173},
  year={2026}
}

@article{opsd,
  title={Self-Distilled Reasoner: On-Policy Self-Distillation for Large Language Models},
  author={Zhao, Siyan and Xie, Zhihui and Liu, Mengchen and Huang, Jing and Pang, Guan and Chen, Feiyu and Grover, Aditya},
  journal={arXiv preprint arXiv:2601.18734},
  year={2026}
}

@article{emotionreasoner,
  title={EmotionReasoner: Emotion-Explanation-Oriented Reinforcement Learning for Explainable Multimodal Emotion Recognition},
  author={Zhao, Sicheng and Zhang, Hongrui and Wu, Daiqing and Huang, Duo and Hong, Weicheng and Sun, Jianyang and Chen, Yifeng and Zhou, Yu and Ding, Guiguang},
  journal={TAFFC},
  year={2026},
}

@article{ppo,
  title={Proximal policy optimization algorithms},
  author={Schulman, John and Wolski, Filip and Dhariwal, Prafulla and Radford, Alec and Klimov, Oleg},
  journal={arXiv preprint arXiv:1707.06347},
  year={2017}
}

@article{does,
  title={Does Reinforcement Learning Really Incentivize Reasoning Capacity in LLMs beyond the Base Model?},
  author={Chen, Zhiqi and Lu, Rui and Zhao, Andrew and Wang, Zhaokai and Yue, Yang and Song, Shiji and Huang, Gao},
  journal={NeurIPS},
  volume={38},
  pages={57654--57689},
  year={2025}
}

@book{minsky_emotion,
  title={Society of Mind},
  author={Minsky, M},
  year={1986},
  publisher={Simon and Schuster}
}

@inproceedings{iclr2026mvei,
    title={Customizing Visual Emotion Evaluation for MLLMs: An Open-vocabulary, Multifaceted, and Scalable Approach},
    author={Daiqing Wu and Dongbao Yang and Sicheng Zhao and Can Ma and Yu Zhou},
    booktitle={ICLR},
    year={2026},
}

\vfill

\end{document}